\documentclass[10pt,twocolumn,letterpaper]{article}

\usepackage[pagenumbers]{cvpr}
\usepackage{url}
\usepackage{graphicx}
\usepackage{booktabs}
\usepackage{amsmath,amssymb,amsfonts,bm}
\usepackage{nicefrac}
\usepackage{multirow}
\usepackage{array}
\usepackage{makecell}
\usepackage{adjustbox}
\usepackage[table]{xcolor}
\usepackage{pifont}
\usepackage{xspace}
\usepackage{microtype}
\usepackage{tabularx}
\usepackage{placeins}
\usepackage{float}
\usepackage{wrapfig}
\usepackage{cuted}
\usepackage{capt-of}

\newcommand{\oursrow}{\rowcolor{cyan!8}}
\newcommand{\x}{\ensuremath{\times}}
\newcommand{\method}{RePix-VFI\xspace}
\newcommand{\grouprow}[2]{\rowcolor{gray!8}\multicolumn{#1}{l}{\textit{#2}}\\}

\definecolor{cvprblue}{rgb}{0.21,0.49,0.74}
\usepackage[pagebackref,breaklinks,colorlinks,allcolors=cvprblue]{hyperref}

\title{RePix-VFI: Re-rendering Observed Pixels for Video Frame Interpolation\\
with Generative Priors from One-Step Diffusion}

\author{Dayeon Kim\\
[0.3em]
KAIST\\
{\tt\small dyk4501@kaist.ac.kr}
\and
Jeonghyeok Do\footnotemark[2]\\
[0.3em]
KAIST\\
{\tt\small ehwjdgur0913@kaist.ac.kr}
\and
Munchurl Kim\footnotemark[2]\\
[0.3em]
KAIST\\
{\tt\small mkimee@kaist.ac.kr}
\and
\small{\url{https://kaist-viclab.github.io/RePix-VFI_site}}
}

\begin{document}
\maketitle

\begin{strip}
    \centering
    \vspace{-1.5cm}
    \includegraphics[width=\linewidth]{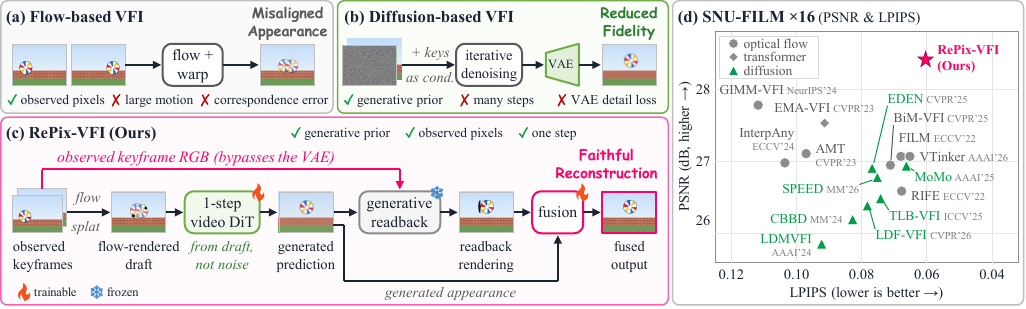}
    \vspace{-0.6cm}
    \captionof{figure}{\textbf{\method{} re-renders observed pixels using a one-step generated video as the correspondence reference.} (a) Flow-based VFI transports observed appearance but can misplace details under inaccurate correspondences. (b) Diffusion-based VFI draws on generative priors but typically requires iterative denoising and can lose observed details in VAE reconstruction. (c) Our RePix-VFI refines a flow-rendered draft in a single forward pass. The completed prediction provides correspondence references for re-rendering original keyframe RGB outside the VAE and supplies complementary appearance for fusion. (d) On SNU-FILM $\times$16, RePix-VFI achieves both the highest PSNR and the lowest LPIPS among the compared methods (upper right is better).}
    \label{fig:teaser}
\end{strip}

{
  \renewcommand{\thefootnote}%
    {\fnsymbol{footnote}}
  \footnotetext[2]{Co-corresponding authors (equal advising).}
}

\begin{abstract}
Multi-frame video frame interpolation (VFI) aims to reconstruct dense video from sparse keyframes, requiring accurate intermediate motion and faithful appearance.
As temporal gaps grow, flow-based methods can retain observed textures but misplace them under ambiguous motion, whereas iterative latent-diffusion models can synthesize plausible content but incur substantial sampling costs and may alter details during reconstruction with a variational autoencoder (VAE).
We present \textbf{RePix-VFI}, named for \textbf{Re}-rendering observed \textbf{Pix}els, a multi-frame VFI framework that couples draft-conditioned one-step video diffusion with reconstruction from observed keyframes.
A pretrained video diffusion model refines a flow-rendered draft of the whole window in a single forward pass. For each target frame, our \textit{generative readback} matches its two bounding keyframes to the generated prediction using an optical-flow estimator. The resulting correspondences guide re-rendering of original keyframe RGB pixels outside the VAE. A lightweight appearance fusion head combines these renderings with generated appearance.
Experiments on benchmark datasets demonstrate significant improvements in reconstruction and perceptual quality. 
On SNU-FILM at $\times$8 and $\times$16, RePix-VFI achieves the highest PSNR and the lowest LPIPS, FloLPIPS, and tOF among the compared methods, exceeding the best evaluated diffusion-based baseline by 1.71 and 1.48\,dB in PSNR, respectively. It also runs 32--42$\times$ faster than the state-of-the-art multi-frame VFI diffusion model.
\end{abstract}    
\section{Introduction}
\label{sec:intro}

Video frame interpolation (VFI) reconstructs the intermediate frames between observed keyframes, increasing the temporal resolution of a video. It is a fundamental low-level vision task, widely used for slow-motion synthesis~\cite{superslomo2018}, frame-rate up-conversion, and video compression~\cite{vcii2018}, but remains challenging because motion and appearance at unobserved times must be inferred from the available frames. Multi-frame VFI goes beyond a single frame pair, reconstructing a dense video window jointly from several sparse keyframes. Accurate reconstruction requires placing content correctly at intermediate times while preserving observed appearance. As the temporal gap grows, large displacements, nonlinear motion, and occlusions make intermediate motion and visibility ambiguous.

To reconstruct intermediate frames while retaining observed appearance, flow-based VFI methods estimate intermediate motion and warp observed pixels or features accordingly~\cite{rife2022,softsplat2020, film2022,amt2023,gimmvfi2024,bimvfi2025}. As illustrated in Fig.~\ref{fig:teaser}(a), this provides direct access to keyframe textures, but inaccurate correspondences can misplace detail and produce ghosting under large gaps. Moreover, pixel transport alone cannot reconstruct content absent from all observations.

Diffusion-based VFI offers complementary information through learned priors for intermediate appearance or motion~\cite{ldmvfi2024,momo2025,hfd2025}. These priors can support plausible predictions when motion is ambiguous or observations are incomplete. Video diffusion models extend these priors to multi-frame VFI, jointly predicting a whole window from multiple keyframes~\cite{ldfvfi2026}.
However, as shown in Fig.~\ref{fig:teaser}(b), iterative latent-diffusion models incur substantial sampling costs through repeated network evaluations, and reconstructing frames with a variational autoencoder (VAE) can alter details already available in the keyframes.

Prior methods use interpolated estimates to refine motion or match input textures, or derive continuous trajectories from generated latent anchors~\cite{abme2021,vtinker2026,rdvfi2026}.
We instead use the RGB predictions from one-step video generation as references for matching the observed keyframes at each target time. This separates the reference used to estimate correspondences from the observed RGB transported by the renderer. Our controlled comparison shows that a generated prediction can provide a better rendering reference even when its RGB PSNR is lower than that of the flow-rendered draft.

Based on this observation, we present \textbf{RePix-VFI}, named for \textbf{Re}-rendering observed \textbf{Pix}els, a multi-frame VFI framework that couples draft-conditioned one-step video diffusion with reconstruction from observed keyframes (Fig.~\ref{fig:teaser}(c)).
A flow-based renderer first constructs a dense draft from the observations, providing an initial estimate of intermediate motion and appearance. Starting from the encoded flow-rendered draft, a pretrained video diffusion transformer, fine-tuned as a one-step generator, predicts a latent correction in a single forward pass per window. For each intermediate frame, our \textit{generative readback} uses a frozen optical-flow estimator to match its two bounding keyframes to the generated prediction.
The resulting correspondences guide re-rendering of original keyframe RGB along a path that bypasses the VAE. A lightweight appearance fusion head combines these renderings with the generated predictions.

Experiments on SNU-FILM and DAVIS demonstrate the effectiveness of the complete reconstruction framework.
Under the SNU-FILM entire-video protocol at \x8 and \x16, RePix-VFI achieves the highest PSNR and the lowest LPIPS, FloLPIPS, and tOF among the compared methods, exceeding the best evaluated diffusion-based baseline by 1.71 and 1.48\,dB in PSNR, respectively. It also attains the highest PSNR and lowest LPIPS on DAVIS at \x4--\x16.
Controlled comparisons support the design of generative readback. With the flow estimator and renderer fixed, replacing the flow-rendered draft with the generated prediction as the correspondence reference improves rendering PSNR by 0.45--0.72\,dB. The full model also lowers LPIPS relative to Fusion w/o readback at all three factors, and fusion exceeds the better of the re-rendering and the generated prediction by 0.39--1.46\,dB. Under the same 49-frame protocol, end-to-end inference is 32--42$\times$ faster than LDF-VFI~\cite{ldfvfi2026}, the state-of-the-art multi-frame VFI diffusion model.

Our contributions are summarized as follows:
\begin{itemize}
    \item We propose \textbf{RePix-VFI}, a multi-frame VFI framework that adapts a pretrained video diffusion transformer for draft-conditioned one-step prediction and reuses its RGB outputs for correspondence estimation and appearance fusion.
 
    \item We introduce \textit{generative readback}, which uses the generated predictions at intermediate times as correspondence references for a frozen optical-flow estimator. The resulting correspondences guide re-rendering of original keyframe RGB outside the VAE, followed by lightweight appearance fusion with the generated predictions.

    \item Experiments demonstrate superior reconstruction and perceptual performance, together with improved temporal consistency on SNU-FILM and DAVIS. Controlled ablations examine the roles of generated correspondence references and observation rendering in the complete reconstruction.
\end{itemize}
\section{Related Work}
\label{sec:related}

\subsection{Video Frame Interpolation (VFI)}

\noindent\textbf{Flow-based video frame interpolation.}
Flow-based VFI methods synthesize intermediate frames by warping or splatting observed pixels and features~\cite{superslomo2018,softsplat2020}. Recent works improve motion estimation and reconstruction across scales and interpolation times~\cite{rife2022, film2022, emavfi2023}.
GIMM-VFI~\cite{gimmvfi2024} uses coordinate-based implicit motion modeling, while BiM-VFI~\cite{bimvfi2025} introduces bidirectional motion descriptors for non-uniform motion.
Some methods match the inputs against intermediate estimates: ABME~\cite{abme2021} refines motion using an interpolated anchor, while VTinker~\cite{vtinker2026} maps input texture features onto a proxy derived from flow-warped inputs. 

\noindent\textbf{Diffusion-based video frame interpolation.}
Diffusion-based VFI methods model intermediate appearance or motion using generative priors. LDMVFI~\cite{ldmvfi2024} performs conditional latent diffusion with a decoder that predicts deformable kernels for sampling, whereas MoMo~\cite{momo2025} and HFD~\cite{hfd2025} generate optical flows for subsequent frame synthesis. 
For multi-frame VFI, LDF-VFI~\cite{ldfvfi2026} jointly synthesizes temporal chunks and connects them autoregressively, and conditions its VAE decoder on features from the observed video.
SpeedVFI~\cite{speedvfi2026} distills multi-frame interpolation into one-step generation from noise latents conditioned on encoded keyframes, and SPEED~\cite{speed2026} generates each intermediate frame with one-step diffusion in pixel space, avoiding latent encoding and decoding.

\noindent\textbf{Generation and observation reuse.}
Existing methods connect motion estimation, generation, and observation reuse at different stages.
MoG~\cite{mog2025} injects backward-warped input latents and features to constrain generated motion. The training-free SNM-VFI~\cite{snmvfi2026} initializes and iteratively guides video diffusion with flow-derived drafts, then blends flow-based and generated frames with confidence from forward--backward flow consistency. Both use motion estimated from the observations to guide generation.
Other methods exploit generated intermediate representations: MADiff~\cite{madiff2024} updates motion hints from intermediate predictions during iterative sampling and uses them for conditional generation and input-feature alignment within its decoder. RDVFI~\cite{rdvfi2026} generates low-resolution sparse latent keyframes with one-step diffusion, estimates continuous pixel trajectories, and synthesizes target frames from warped inputs and generated latent features.  

Distinguished from previous methods, our RePix-VFI couples draft-conditioned one-step video diffusion with reconstruction from observed keyframes. A single generator evaluation refines the encoded flow-rendered draft into a dense RGB prediction. Whereas RDVFI~\cite{rdvfi2026} derives trajectories from sparse latent anchors, our \textit{generative readback} matches observed keyframes directly to each generated target-time RGB prediction using a frozen optical-flow estimator. These correspondences transport original keyframe RGB outside the VAE, while the generated prediction remains an appearance candidate for fusion.

\subsection{One-step Diffusion for Restoration}
One-step restoration methods adapt pretrained diffusion priors for efficient image and video reconstruction ~\cite{speedvfi2026,rdvfi2026}. OSEDiff~\cite{osediff2024} starts from the encoded degraded image instead of noise and regularizes the output with variational score distillation (VSD). DOVE~\cite{dove2025} adapts a pretrained video diffusion model through latent-space training followed by pixel-space refinement, while SeedVR2~\cite{seedvr22025} combines progressive distillation with adversarial post-training. PS-SR~\cite{pssr2026} trains a source-initialized base model in latent and pixel space, then supplements lightweight model evaluation with iterative refinement. We adopt latent-to-pixel training for draft-conditioned video generation and connect its completed predictions to correspondence estimation and reconstruction from observed keyframes.
\begin{figure*}[t]
  \centering
  \includegraphics[width=\textwidth]{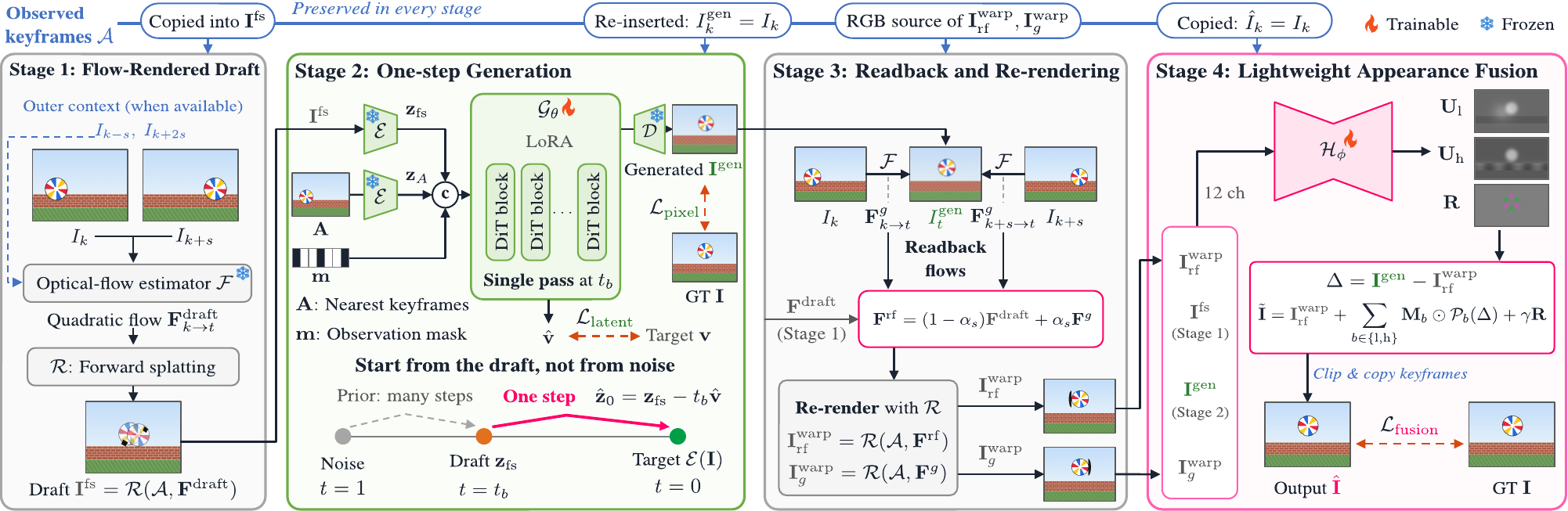}
  \vspace{-0.6cm}
  \caption{\textbf{Overview of \method.}
  (i) \textbf{Stage 1}: A flow-based renderer constructs a draft from observed keyframes.
  (ii) \textbf{Stage 2}: A generator refines the draft latent in one forward pass, conditioned on the observations and their temporal mask.
  (iii) \textbf{Stage 3}: Mixing draft and readback flows yields $\mathbf{I}_{\mathrm{rf}}^{\mathrm{warp}}$, while readback flows alone yield $\mathbf{I}_{g}^{\mathrm{warp}}$.
  (iv) \textbf{Stage 4}: A lightweight appearance fusion head combines four RGB inputs to predict spatial mixing between $\mathbf{I}_{\mathrm{rf}}^{\mathrm{warp}}$ and $\mathbf{I}^{\mathrm{gen}}$, together with a residual correction.}
  \label{fig:framework}
  \vspace{-0.3cm}
\end{figure*}

\section{RePix-VFI}
\label{sec:method}

Fig.~\ref{fig:framework} illustrates the proposed four-stage pipeline. A flow-rendered draft provides an initial estimate of intermediate motion and appearance. A pretrained video diffusion transformer then refines the draft's latent representation in a single forward pass per window. The decoded target frames connect generation to observation reconstruction: each prediction serves as both an RGB reference for estimating keyframe-to-target correspondences and an appearance candidate for fusion. The rendering path thus uses generated predictions for matching while drawing its RGB pixels from the observations, outside the VAE. A lightweight appearance fusion head combines these renderings with generated appearance.

Let $\mathbf{I}=\{I_t\}_{t=0}^{T-1}$ denote a target ground-truth video of $T$ frames. For an interpolation factor $s$, the observed keyframes are $\mathcal{A}=\{I_k\}_{k\in\mathcal{K}_s}$, where $\mathcal{K}_s$ indexes frames sampled at intervals of $s$. The goal is to reconstruct the missing frames indexed by $\mathcal{M}_s=\{0,\ldots,T-1\}\setminus\mathcal{K}_s$, which we refer to as target frames.

\subsection{Flow-Rendered Draft}
\label{sec:draft}

\noindent\textbf{Optical flow estimation.}
For a target time $t\in \mathcal{M}_s$, let $\{k, k+s\}\subset \mathcal{K}_s$ be its bounding keyframe indices corresponding to two consecutive keyframes.
In Stage 1 of Fig.~\ref{fig:framework}, a frozen optical-flow estimator $\mathcal{F}$~\cite{raft2020} computes flows from $I_k$ to its adjacent observed keyframes, $I_{k\pm s}$.
Following quadratic motion interpolation~\cite{qvi2019}, the flow from the left keyframe is given by
\begin{equation}
\begin{aligned}
\mathbf{F}^{\mathrm{draft}}_{k\to t}
=&{} \frac{\tau}{2}
\left(
\mathbf{F}_{k\to k+s}
-
\mathbf{F}_{k\to k-s}
\right) \\
&+ \frac{\tau^2}{2}
\left(
\mathbf{F}_{k\to k+s}
+
\mathbf{F}_{k\to k-s}
\right),
\end{aligned}
\label{eq:draft_motion}
\end{equation}
where $\tau=(t-k)/s\in(0,1)$ is the normalized interpolation time.
The right-keyframe field $\mathbf{F}^{\mathrm{draft}}_{k+s\to t}$ is estimated symmetrically using flows from $I_{k+s}$ to $I_k$ and $I_{k+2s}$.
When the required outer keyframe is unavailable or the quadratic fit is invalid, we use linear motion from the corresponding bounding keyframe.

\noindent\textbf{Draft rendering.}
Let $\mathbf{F}^{\mathrm{draft}}$ stack the left- and right-keyframe intermediate flows for all target frames.
We construct a draft $\mathbf{I}^{\mathrm{fs}}$ via forward splatting to serve as the starting point for one-step diffusion generation:
\begin{equation}
\mathbf{I}^{\mathrm{fs}}
=
\mathcal{R}(\mathcal{A},\mathbf{F}^{\mathrm{draft}}),
\label{eq:draft_render}
\end{equation}
where $\mathcal{R}$ denotes the rendering operator.
At each target time $t$, $\mathcal{R}$ bilinearly forward-splats the two bounding keyframes along their displacement fields~\cite{softsplat2020}.
The splatted RGB contributions are weighted by $(1-\tau,\tau)$ and normalized by the corresponding weighted splat mass.
Pixels receiving no contribution are filled with the corresponding values from the unwarped linear blend $(1-\tau)I_k+\tau I_{k+s}$.
Observed keyframes are copied unchanged into $\mathbf{I}^{\mathrm{fs}}$.

\subsection{Draft-Conditioned One-Step Generation}
\label{sec:generator}

\noindent\textbf{Conditioning.}
As shown in Stage 2 of Fig.~\ref{fig:framework}, the encoded flow-rendered draft serves as the initial latent state for a pretrained video diffusion transformer $\mathcal{G}_{\theta}$~\cite{wan2025}.
We adapt the transformer for deterministic one-step refinement, with additional conditioning from the observed keyframes.
Let $\mathcal{E}$ and $\mathcal{D}$ denote the frozen VAE encoder and decoder, respectively.
We construct $\mathbf{A}$ by copying the temporally nearest observation to each target index and encode the two videos as $\mathbf{z}_{\mathrm{fs}}=\mathcal{E}(\mathbf{I}^{\mathrm{fs}})$ and $\mathbf{z}_A=\mathcal{E}(\mathbf{A})$.
Inspired by the channel-wise video conditioning scheme~\cite{ldfvfi2026}, we concatenate $\mathbf{z}_{\mathrm{fs}}$, $\mathbf{z}_A$, and an observation mask $\mathbf{m}$ along the channel dimension.
The mask encodes the observed frame positions in a layout compatible with the VAE's temporal compression, distinguishing the observed keyframes from the copied ones for the target frames.

\noindent\textbf{One-step refinement.}
We use $\mathbf{z}_{\mathrm{fs}}$ directly, without adding diffusion noise, and evaluate $\mathcal{G}_{\theta}$ at a fixed normalized timestep $t_b$. With the pretrained rectified-flow velocity parameterization~\cite{rectifiedflow2023,flowmatching2023}, the predicted velocity defines a single latent correction toward the refined latent $\hat{\mathbf{z}}_0$ as:
\begin{equation}
\begin{aligned}
\hat{\mathbf{v}}
=
\mathcal{G}_{\theta}
\big([\mathbf{z}_{\mathrm{fs}},\mathbf{z}_A,\mathbf{m}],t_b\big), \qquad  \hat{\mathbf{z}}_0 = \mathbf{z}_{\mathrm{fs}}-t_b\hat{\mathbf{v}}
\end{aligned}
\label{eq:onestep}
\end{equation}
Then, we obtain $\tilde{\mathbf{I}}^{\mathrm{gen}} = \mathcal{D}(\hat{\mathbf{z}}_0)$ after decoding. In $\tilde{\mathbf{I}}^{\mathrm{gen}}$, we replace the generated frames at the observed frame positions with the corresponding observed frames, yielding $\mathbf{I}^{\mathrm{gen}}=\{I_t^{\mathrm{gen}}\}_{t=0}^{T-1}$ so that $I_k^{\mathrm{gen}}=I_k$ for all $k\in\mathcal{K}_s$. Note that $\mathcal{G}_{\theta}$ shares the same parameters across interpolation factors.

\noindent\textbf{Generation objective.}
We train LoRA adapters~\cite{lora2022} in the attention and feed-forward layers of $\mathcal{G}_{\theta}$, together with the patch embedding and output projection; the remaining backbone parameters and the VAE are frozen.
Newly introduced patch-embedding weights are zero-initialized.
Building on source-initialized latent refinement and latent-to-pixel adaptation~\cite{pssr2026}, we train the same generator in two successive stages.
The latent stage supervises the draft-to-target correction using the target velocity
\begin{equation}
\mathbf{v}
=
\big({\mathbf{z}_{\mathrm{fs}}-\mathcal{E}(\mathbf{I})}\big)/{t_b},
\label{eq:vstar}
\end{equation}
where the latent-stage objective is
\begin{equation}
\mathcal{L}_{\mathrm{latent}}
=
\mathcal{L}_{\mathrm{mse}}
(\hat{\mathbf{v}},\mathbf{v})
+
\lambda_{\mathrm{vsd}}\mathcal{L}_{\mathrm{vsd}}
+
\lambda_{\mathrm{adv}}\mathcal{L}_{\mathrm{adv}},
\label{eq:lat}
\end{equation}
where $\lambda_{\mathrm{vsd}}=1$ and $\lambda_{\mathrm{adv}}=0.1$ are empirically set for all our experiments. 
The VSD loss compares predictions from a frozen reference score model and a learned auxiliary score model on re-noised generated latents~\cite{vsd2023,dmd2024}. The adversarial loss uses a latent-space discriminator trained with a hinge loss. 
In the pixel stage, we subsequently refine the same generator using
\begin{equation}
\mathcal{L}_{\mathrm{pixel}}
=
\mathcal{L}^{\mathcal{M}_s}_{\mathrm{mse}}
(\mathbf{I}^{\mathrm{gen}},\mathbf{I}) \\
+
\lambda_{\mathrm{lpips}}
\mathcal{L}^{\mathcal{M}_s}_{\mathrm{lpips}}
(\mathbf{I}^{\mathrm{gen}},\mathbf{I}),
\label{eq:pix}
\end{equation}
where $\lambda_{\mathrm{lpips}}=2$ and the superscript $\mathcal{M}_s$ restricts both loss functions to missing frames that we target.
The two training stages supervise the draft-to-target latent correction and its decoded appearance, respectively.

\subsection{Readback and Re-rendering}
\label{sec:rendering}

\noindent\textbf{Generative readback.}
After decoding, each target prediction $I_t^{\mathrm{gen}}$ provides an explicit RGB reference at the unobserved time. 
In Stage 3 of Fig.~\ref{fig:framework}, we reuse the frozen optical-flow estimator $\mathcal{F}$ to match both bounding keyframes directly to this prediction:
\begin{equation}
\mathbf{F}^{g}_{k\to t}
= \mathcal{F}(I_k,I_t^{\mathrm{gen}}), \quad
\mathbf{F}^{g}_{k+s\to t}
= \mathcal{F}(I_{k+s},I_t^{\mathrm{gen}}).
\label{eq:readback_flow}
\end{equation}
Whereas the draft estimates target-time motion from correspondences between observations, readback estimates correspondences against the generated target-time RGB. This uses the flow estimator without VFI-specific correspondence training. Stacking these fields gives $\mathbf{F}^{g}$, which we blend with the draft flows to control the readback update:
\begin{equation}
\mathbf{F}^{\mathrm{rf}}
=
\big(1-\alpha_s\big) \mathbf{F}^{\mathrm{draft}}
+
\alpha_s \mathbf{F}^{g},
\label{eq:flow_update}
\end{equation}
where $\alpha_s\in[0,1]$ is calibrated separately for each interpolation factor $s$.
Grid search on designated fusion-training windows selects $\alpha_s=0.8, 0.9, 1.0$ for $s=4, 8, 16$.

\noindent\textbf{Re-rendering.}
We apply the rendering operator from Sec.~\ref{sec:draft} to the refined and readback-only flows:
\begin{equation}
\mathbf{I}_{\mathrm{rf}}^{\mathrm{warp}}
=
\mathcal{R}(\mathcal{A},\mathbf{F}^{\mathrm{rf}}),
\qquad
\mathbf{I}_{g}^{\mathrm{warp}}
=
\mathcal{R}(\mathcal{A},\mathbf{F}^{g}).
\label{eq:rerender}
\end{equation}
Both renderings draw their RGB values from the original keyframes, while the generated prediction guides their transport through the estimated correspondences. We use $\mathbf{I}_{\mathrm{rf}}^{\mathrm{warp}}$ as the reconstruction base and $\mathbf{I}_{g}^{\mathrm{warp}}$ as an auxiliary fusion input; they coincide when $\alpha_s=1$.

\subsection{Lightweight Appearance Fusion}
\label{sec:fusion}
Stage 4 of Fig.~\ref{fig:framework} combines generated appearance $\mathbf{I}^{\mathrm{gen}}$ with the observation rendering $\mathbf{I}_{\mathrm{rf}}^{\mathrm{warp}}$ to produce the final reconstruction.
A framewise 2D U-Net $\mathcal{H}_{\phi}$~\cite{unet2015} predicts their spatial mixing and a residual correction, using the draft $\mathbf{I}^{\mathrm{fs}}$ and readback-only rendering $\mathbf{I}_{g}^{\mathrm{warp}}$ as additional conditioning.
The four RGB inputs are concatenated into 12 channels, from which the head predicts two mixing logits, $\mathbf{U}_{\mathrm{l}}$ and $\mathbf{U}_{\mathrm{h}}$, and a 3-channel residual $\mathbf{R}$:
\begin{equation}
[\mathbf{U}_{\mathrm{l}},\mathbf{U}_{\mathrm{h}}, \mathbf{R}]
=
\mathcal{H}_\phi
\big([\mathbf{I}_{\mathrm{rf}}^{\mathrm{warp}},\mathbf{I}^{\mathrm{fs}},\mathbf{I}^{\mathrm{gen}},\mathbf{I}_{g}^{\mathrm{warp}}]\big).
\label{eq:fusion_head}
\end{equation}
Let $\Delta=\mathbf{I}^{\mathrm{gen}}-\mathbf{I}_{\mathrm{rf}}^{\mathrm{warp}}$ denote the difference between the two appearance candidates.
Separate spatial maps modulate its low- and high-frequency components, while the residual provides an additional RGB correction:
\begin{equation}
\tilde{\mathbf{I}}
=
\mathbf{I}_{\mathrm{rf}}^{\mathrm{warp}}
+
\sum\nolimits_{b\in\{\mathrm{l},\mathrm{h}\}}
\mathbf{M}_b\odot\mathcal{P}_b(\Delta)
+ 
\gamma\mathbf{R},
\label{eq:fusion_output}
\end{equation}
where $\mathbf{M}_b=\mathrm{sigmoid}(\mathbf{U}_b)$. 
We empirically set $\gamma=0.03$, allowing only small residual corrections beyond the adaptive mixing of re-rendered observations and generated appearance. The head includes calibrated logit offsets and a bounded residual transformation.
Here, $\mathcal{P}_{\mathrm{l}}$ is a Gaussian low-pass filter with a standard deviation of 2 pixels and $\mathcal{P}_{\mathrm{h}}=\mathcal{I}-\mathcal{P}_{\mathrm{l}}$, where $\mathcal{I}$ is the identity operator.
At inference, we clip $\tilde{\mathbf{I}}$ to $[0,1]$ and copy the observed keyframes into the result to obtain the final output $\hat{\mathbf{I}}$, ensuring $\hat{I}_k=I_k$ for all $k\in\mathcal{K}_s$.

\noindent\textbf{Fusion objective.}
With the generator $\mathcal{G}_{\theta}$ frozen, we optimize the appearance fusion head $\mathcal{H}_\phi$ using draft-relative reconstruction and perceptual losses:
\begin{equation}
\mathcal{L}_{\mathrm{fusion}}
=
\mathcal{L}^{\mathrm{rel},\mathcal{M}_s}_{\mathrm{mse}}
(\tilde{\mathbf{I}},\mathbf{I}) \\
+
\lambda_{\mathrm{lpips}}
\mathcal{L}^{\mathrm{rel},\mathcal{M}_s}_{\mathrm{lpips}}
(\hat{\mathbf{I}},\mathbf{I}),
\label{eq:fusion_loss}
\end{equation}
where $\mathrm{rel}$ denotes normalization by the corresponding draft error, and the perceptual objective includes a weighted component with temporal-distance. More details are provided in the \textit{Appendix}.

\section{Experiments}
\label{sec:experiment}

\begin{table*}[!t]
\centering
\caption{\textbf{Quantitative comparison on SNU-FILM under the entire-video protocol.}
Frame metrics are evaluated on missing frames from 31 clips; tOF uses all adjacent frame pairs, including observed-frame transitions. Scores are averaged equally over clips.
\textbf{Bold} and \underline{underline} indicate the best and second-best results. NFE is reported per synthesized frame, except per chunk for LDF-VFI and per window for RePix-VFI. Midpoint-only methods are applied recursively.}
\vspace{-0.2cm}
\label{tab:snufilm}
\small
\setlength{\tabcolsep}{2.5pt}
\renewcommand{\arraystretch}{1.1}
\resizebox{\textwidth}{!}{%
\begin{tabular}{llcccccccccccccccc}
\toprule
\multirow{2}{*}{\textbf{Method}} & \multirow{2}{*}{\textbf{Venue}} & \multirow{2}{*}{\textbf{NFE}} & \multicolumn{5}{c}{\textbf{SNU-FILM \x4}} & \multicolumn{5}{c}{\textbf{SNU-FILM \x8}} & \multicolumn{5}{c}{\textbf{SNU-FILM \x16}} \\
\cmidrule(lr){4-8}\cmidrule(lr){9-13}\cmidrule(lr){14-18}
 & &  & PSNR$\uparrow$ & SSIM$\uparrow$ & LPIPS$\downarrow$ & FloLPIPS$\downarrow$ & tOF$\downarrow$ & PSNR$\uparrow$ & SSIM$\uparrow$ & LPIPS$\downarrow$ & FloLPIPS$\downarrow$ & tOF$\downarrow$ & PSNR$\uparrow$ & SSIM$\uparrow$ & LPIPS$\downarrow$ & FloLPIPS$\downarrow$ & tOF$\downarrow$ \\
\midrule

\grouprow{18}{Non-diffusion}
FILM~\cite{film2022} & ECCV'22 & 1 & 36.62 & 0.9815 & 0.0186 & \textbf{0.0300} & 0.2856 & 31.91 & 0.9501 & 0.0344 & \underline{0.0565} & 0.4129 & 27.07 & 0.8879 & 0.0680 & 0.1094 & 0.6814 \\
RIFE~\cite{rife2022} & ECCV'22 & 1 & 35.72 & 0.9790 & 0.0211 & 0.0340 & 0.3248 & 31.19 & 0.9479 & 0.0364 & 0.0618 & 0.4601 & 26.59 & 0.8854 & 0.0678 & 0.1114 & 0.7617 \\
AMT~\cite{amt2023} & CVPR'23 & 1 & 36.73 & 0.9823 & 0.0291 & 0.0470 & 0.3766 & 32.04 & 0.9525 & 0.0495 & 0.0855 & 0.5086 & 27.11 & 0.8926 & 0.0971 & 0.1659 & 0.7852 \\
EMA-VFI~\cite{emavfi2023} & CVPR'23 & 1 & 36.63 & 0.9827 & 0.0296 & 0.0492 & 0.2924 & 32.27 & 0.9543 & 0.0496 & 0.0840 & 0.4073 & 27.53 & 0.8962 & 0.0914 & 0.1537 & 0.6055 \\
GIMM-VFI~\cite{gimmvfi2024} & NeurIPS'24 & 1 & \underline{36.88} & \underline{0.9828} & 0.0336 & 0.0527 & \underline{0.2811} & \underline{32.45} & \underline{0.9554} & 0.0607 & 0.0970 & \underline{0.3947} & \underline{27.78} & \textbf{0.9016} & 0.1118 & 0.1762 & \underline{0.5674} \\
InterpAny~\cite{interpany2024} & ECCV'24 & 2 & 36.24 & 0.9809 & 0.0278 & 0.0477 & 0.3246 & 31.60 & 0.9496 & 0.0510 & 0.0926 & 0.4807 & 26.98 & 0.8898 & 0.1036 & 0.1717 & 0.7592 \\
BiM-VFI~\cite{bimvfi2025} & CVPR'25 & 1 & 36.43 & 0.9812 & 0.0230 & 0.0371 & 0.3265 & 31.69 & 0.9493 & 0.0394 & 0.0650 & 0.4713 & 26.95 & 0.8886 & 0.0712 & 0.1149 & 0.6522 \\
VTinker~\cite{vtinker2026} & AAAI'26 & 1 & 36.04 & 0.9805 & 0.0186 & \underline{0.0306} & 0.2902 & 31.67 & 0.9492 & 0.0340 & 0.0569 & 0.4141 & 27.07 & 0.8890 & \underline{0.0652} & \underline{0.1072} & 0.6186 \\
\midrule
\grouprow{18}{Diffusion-based}
LDMVFI~\cite{ldmvfi2024} & AAAI'24 & 200 & 34.79 & 0.9746 & 0.0249 & 0.0450 & 0.3408 & 30.15 & 0.9364 & 0.0480 & 0.0877 & 0.5340 & 25.86 & 0.8727 & 0.0923 & 0.1521 & 0.8668 \\
CBBD~\cite{cbbd2024} & MM'24 & 50 & 35.95 & 0.9799 & 0.0194 & 0.0339 & 0.3000 & 31.02 & 0.9451 & 0.0390 & 0.0721 & 0.5148 & 26.20 & 0.8779 & 0.0828 & 0.1371 & 0.9153 \\
MoMo~\cite{momo2025} & AAAI'25 & 8 & 36.20 & 0.9801 & \textbf{0.0178} & \textbf{0.0300} & 0.3011 & 31.60 & 0.9480 & \underline{0.0336} & 0.0577 & 0.4207 & 26.94 & 0.8869 & 0.0663 & 0.1110 & 0.6247 \\
EDEN~\cite{eden2025} & CVPR'25 & 2 & 35.44 & 0.9780 & 0.0203 & 0.0372 & 0.2913 & 31.14 & 0.9462 & 0.0392 & 0.0708 & 0.4185 & 26.91 & 0.8885 & 0.0768 & 0.1276 & 0.6337 \\
TLB-VFI~\cite{tlbvfi2025} & ICCV'25 & 10 & 35.94 & 0.9802 & \underline{0.0182} & 0.0316 & 0.3092 & 30.96 & 0.9455 & 0.0367 & 0.0686 & 0.5193 & 26.49 & 0.8820 & 0.0742 & 0.1152 & 0.7655 \\
SPEED~\cite{speed2026} & MM'26 & 1 & 34.37 & 0.9721 & 0.0257 & 0.0391 & 0.2888 & 30.70 & 0.9420 & 0.0430 & 0.0703 & 0.4086 & 26.78 & 0.8857 & 0.0752 & 0.1196 & 0.5770 \\
LDF-VFI~\cite{ldfvfi2026} & CVPR'26 & 16 & 33.47 & 0.9670 & 0.0320 & 0.0428 & 0.3133 & 30.79 & 0.9442 & 0.0452 & 0.0634 & 0.4099 & 26.39 & 0.8794 & 0.0783 & 0.1160 & 0.5911 \\
\oursrow
\textbf{\method} & -- & 1 & \textbf{36.98} & \textbf{0.9842} & 0.0198 & 0.0310 & \textbf{0.2474} & \textbf{33.31} & \textbf{0.9628} & \textbf{0.0322} & \textbf{0.0493} & \textbf{0.3490} & \textbf{28.42} & \underline{0.9014} & \textbf{0.0604} & \textbf{0.0871} & \textbf{0.5400} \\
\bottomrule
\end{tabular}}
\vspace{-0.3cm}
\end{table*}

\begin{table}[!t]
\centering
\caption{\textbf{Quantitative comparison on DAVIS 2016.}
PSNR and LPIPS are evaluated on missing frames from all 50 sequences at their native resolutions and averaged equally over sequences.}
\vspace{-0.2cm}
\label{tab:davis}
\scriptsize
\setlength{\tabcolsep}{3pt}
\renewcommand{\arraystretch}{1.1}
\resizebox{\columnwidth}{!}{%
\begin{tabular}{lcccccc}
\toprule
\multirow{2}{*}{\textbf{Method}} & \multicolumn{2}{c}{\textbf{\x4}} & \multicolumn{2}{c}{\textbf{\x8}} & \multicolumn{2}{c}{\textbf{\x16}} \\
\cmidrule(lr){2-3}\cmidrule(lr){4-5}\cmidrule(lr){6-7}
 & PSNR$\uparrow$ & LPIPS$\downarrow$ & PSNR$\uparrow$ & LPIPS$\downarrow$ & PSNR$\uparrow$ & LPIPS$\downarrow$ \\
\midrule
GIMM-VFI~\cite{gimmvfi2024} & \underline{23.81} & 0.2737 & \underline{20.59} & 0.3903 & 18.22 & 0.5046 \\
BiM-VFI~\cite{bimvfi2025} & 22.86 & \underline{0.1643} & 19.71 & \underline{0.2451} & 17.40 & \underline{0.3413} \\
VTinker~\cite{vtinker2026} & 22.71 & 0.1742 & 19.47 & 0.2680 & 17.39 & 0.3575 \\
MoMo~\cite{momo2025} & 22.50 & 0.1718 & 19.33 & 0.2638 & 17.26 & 0.3523 \\
EDEN~\cite{eden2025} & 22.34 & 0.3260 & 20.06 & 0.4239 & \underline{18.38} & 0.4940 \\
TLB-VFI~\cite{tlbvfi2025} & 20.62 & 0.2367 & 18.30 & 0.3097 & 17.14 & 0.3630 \\
LDF-VFI~\cite{ldfvfi2026} & 22.74 & 0.2444 & 20.11 & 0.3055 & 17.71 & 0.3974 \\
\oursrow
\textbf{\method} & \textbf{24.56} & \textbf{0.1422} & \textbf{21.64} & \textbf{0.2099} & \textbf{19.17} & \textbf{0.2962} \\
\bottomrule
\end{tabular}}
\vspace{-0.3cm}
\end{table}

\subsection{Experimental Setup}
\label{sec:setup}

\noindent\textbf{Training data and implementation.}
We fine-tune Wan2.2-TI2V-5B~\cite{wan2025} on LAVIB~\cite{lavib2024} using rank-32 LoRA~\cite{lora2022} and 49-frame, $512\times512$ crops, with keyframe intervals sampled uniformly from 2 to 16 and random temporal phases.
Generator adaptation comprises 5k latent-stage and 30k pixel-stage updates at $t_b=0.599$, while the appearance fusion head is trained for 13.5k updates in total using frozen generator outputs.
Optimization, fusion initialization, and other details are provided in the \textit{Appendix}.

\noindent\textbf{Benchmarks and baselines.}
We evaluate at interpolation factors $s\in\{4,8,16\}$ using the SNU-FILM entire-video protocol~\cite{cain2020,ldfvfi2026} and DAVIS 2016~\cite{davis2016} at native 1080p resolutions.
At each factor, every $s$-th frame is retained as a keyframe, and the remaining frames are reconstructed.
Table~\ref{tab:snufilm} compares our RePix-VFI with 15 VFI baselines using released weights, covering non-diffusion and diffusion-based VFI.
All methods reconstruct the same target frames from the same subsampled videos using their native temporal contexts.

\noindent\textbf{Metrics.}
We measure reconstruction fidelity using PSNR and SSIM~\cite{ssim2004}, perceptual similarity using
AlexNet-LPIPS~\cite{lpips2018}, and motion-aware perceptual quality using FloLPIPS~\cite{flolpips2022}. We additionally evaluate temporal consistency using a RAFT-based implementation of tOF~\cite{tecogan2020,raft2020}. 
PSNR, SSIM, LPIPS, and FloLPIPS are computed on missing frames, whereas tOF uses all adjacent frame pairs, including observed-frame transitions. The metrics are averaged equally over sequences.

\subsection{Quantitative Comparisons}
\label{sec:main}
\noindent\textbf{SNU-FILM.}
As shown in Table~\ref{tab:snufilm}, our RePix-VFI achieves the highest PSNR at all three factors and the lowest LPIPS and FloLPIPS at \x8 and \x16. At these two factors, PSNR improves over GIMM-VFI~\cite{gimmvfi2024} by $0.86,0.64$\,dB and over MoMo~\cite{momo2025}, the strongest evaluated diffusion-based baseline, by $1.71,1.48$\,dB.
For the temporal consistency metric tOF, our RePix-VFI reduces the error relative to the best competing result with large margins of 12.0\%, 11.6\%, and 4.8\% at \x4, \x8, and \x16, respectively.
Together, these results show improved reconstruction accuracy and perceptual quality, accompanied by lower temporal motion discrepancy.

\noindent\textbf{DAVIS.}
Table~\ref{tab:davis} reports the VFI results averaged over all 50 DAVIS 2016 sequences. Our RePix-VFI achieves the highest PSNR and lowest LPIPS at all three interpolation factors, with PSNR gains of 0.75, 1.04, and 0.79\,dB over the best competing results at \x4, \x8, and \x16, respectively.
This extends the reconstruction and perceptual-quality improvement on high-resolution DAVIS sequences.

\subsection{Qualitative Comparisons}
\label{sec:qualitative}

\begin{figure*}[tbp]
  \centering
  \includegraphics[width=\textwidth]{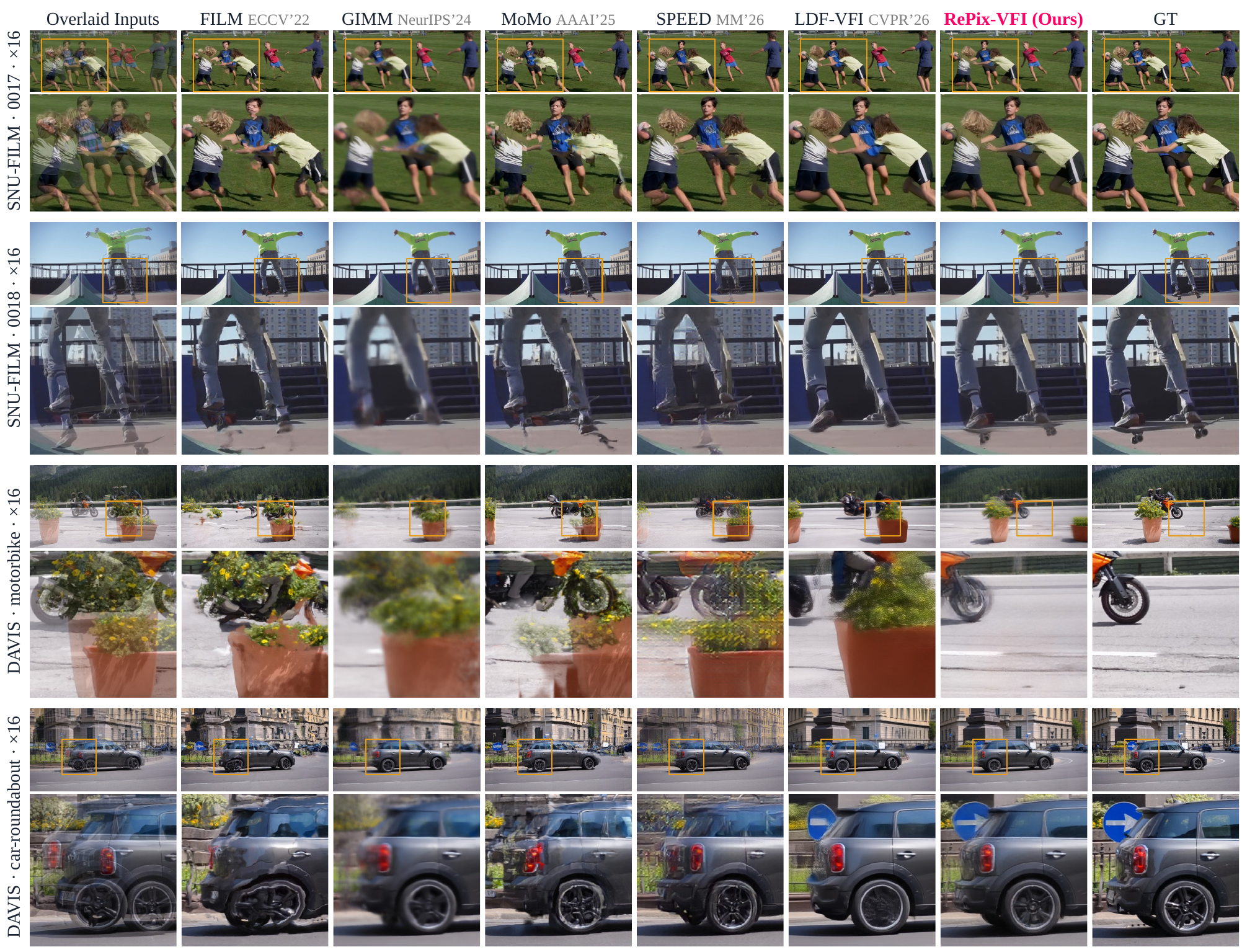}
  \vspace{-0.7cm}
  \caption{\textbf{Qualitative comparisons at $\times$16.} The top two examples are from SNU-FILM and the bottom two from DAVIS, all at the midpoint between observed keyframes ($\tau=8/16$). 
  The first column shows an equal-weight overlay of the bounding observations; subsequent columns show predictions and GT. Boxes indicate the regions enlarged below.}
  \label{fig:qualitative}
  \vspace{-0.3cm}
\end{figure*}

Fig.~\ref{fig:qualitative} compares reconstructed frames at \x16 on SNU-FILM and DAVIS.
In the SNU-FILM examples, our RePix-VFI better preserves the relative positions of overlapping people's legs and the configuration of the skateboarder's legs and board.
Several competing predictions exhibit duplicated contours, distorted structures, or oversmoothed details in these regions.
This advantage in intermediate placement is also evident in the fast-motion DAVIS motorbike example, where our reconstruction more closely matches the GT positions of the motorcycle and foreground planter.
Beyond placement, the car-roundabout example shows more faithful rear-wheel geometry and car-body contours, together with the visible directional arrow on the partially occluded road sign behind the vehicle.
These examples demonstrate the framework's ability to recover intermediate scene configurations while preserving local structure and appearance.

\begin{table}[t]
\centering
\caption{\textbf{Reconstruction ablations on SNU-FILM.}
Scores follow the entire-video protocol.
\textit{$\mathbf{I}^{\mathrm{fs}}$}-reference rendering replaces $\mathbf{I}^{\mathrm{gen}}$ with $\mathbf{I}^{\mathrm{fs}}$ as the correspondence reference under fixed rendering settings. With the generator frozen, \textit{Fusion w/o readback} and \textit{Full reconstruction} are trained with the same training strategy.}
\label{tab:ablation}
\vspace{-0.2cm}
\label{tab:components}
\scriptsize
\setlength{\tabcolsep}{1.5pt}
\renewcommand{\arraystretch}{1.1}
\resizebox{\linewidth}{!}{%
\begin{tabular}{lcccccc}
\toprule
\multirow{2}{*}{\textbf{Output}} & \multicolumn{2}{c}{\textbf{\x4}} & \multicolumn{2}{c}{\textbf{\x8}} & \multicolumn{2}{c}{\textbf{\x16}} \\
\cmidrule(lr){2-3}\cmidrule(lr){4-5}\cmidrule(lr){6-7}
 & PSNR$\uparrow$ & LPIPS$\downarrow$ & PSNR$\uparrow$ & LPIPS$\downarrow$ & PSNR$\uparrow$ & LPIPS$\downarrow$ \\
\midrule
Draft $\mathbf{I}^{\mathrm{fs}}$ & 35.15 & 0.0374 & 31.33 & 0.0598 & 27.12 & 0.0966 \\
Generation $\mathbf{I}^{\mathrm{gen}}$ ($\mathcal{L}_{\mathrm{latent}}$) & 33.24 & 0.0378 & 30.35 & 0.0599 & 26.59 & 0.0965 \\
Generation $\mathbf{I}^{\mathrm{gen}}$ ($+\mathcal{L}_{\mathrm{pixel}}$) & 34.83 & 0.0286 & 32.19 & 0.0402 & 28.03 & 0.0676 \\
$\mathbf{I}^{\mathrm{fs}}$-reference rendering & 35.07 & 0.0358 & 31.21 & 0.0538 & 26.89 & 0.0888 \\
Re-rendering $\mathbf{I}^{\mathrm{warp}}_{\mathrm{rf}}$ & 35.52 & 0.0349 & 31.93 & 0.0519 & 27.46 & 0.0857 \\
Fusion w/o readback & 36.84 & 0.0228 & 33.04 & 0.0359 & 28.22 & 0.0646 \\
\midrule
\oursrow
\textbf{Full reconstruction (Ours)} & \textbf{36.98} & \textbf{0.0198} & \textbf{33.31} & \textbf{0.0322} & \textbf{28.42} & \textbf{0.0604} \\
\bottomrule
\end{tabular}}
\vspace{-0.3cm}
\end{table}

\subsection{Ablation Studies}
\label{sec:ablation}

\noindent\textbf{Generator Adaptation.}
The two \textit{Generation $\mathbf{I}^{\mathrm{gen}}$} rows in \Cref{tab:components} compare generator outputs before and after pixel-space adaptation on SNU-FILM.
This additional adaptation improves PSNR by 1.44--1.84\,dB and lowers LPIPS.
The resulting $\mathbf{I}^{\mathrm{gen}}$ also achieves lower LPIPS than \textit{Draft $\mathbf{I}^{\mathrm{fs}}$} at all three factors, providing an appearance candidate with improved perceptual similarity for subsequent reconstruction.

\noindent\textbf{Correspondence Reference for Re-rendering.}
To assess whether generation also benefits observation rendering, we compare \textit{$\mathbf{I}^{\mathrm{fs}}$-reference rendering} with \textit{Re-rendering $\mathbf{I}_{\mathrm{rf}}^{\mathrm{warp}}$} in \Cref{tab:components}.
These use $\mathbf{I}^{\mathrm{fs}}$ and $\mathbf{I}^{\mathrm{gen}}$ as their respective correspondence references, with the observed RGB, frozen flow estimator, rendering resolution, renderer, and $\alpha_s$ held fixed.
Using the generated reference improves rendering PSNR by 0.45, 0.72, and 0.57\,dB at \x4, \x8, and \x16 and lowers LPIPS. At $\times4$, the generated prediction has lower PSNR than the draft, yet provides a better reference for observation rendering. Thus, RGB PSNR alone does not determine the usefulness of a prediction as a correspondence reference, supporting the use of generated predictions beyond their role as appearance candidates.

\begin{figure*}[t]
  \centering
  \includegraphics[width=0.91\textwidth]{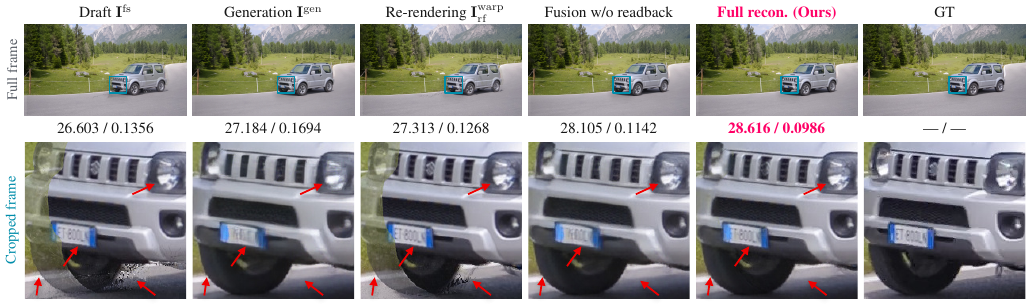}
  \vspace{-0.3cm}
  \caption{\textbf{Complementary roles of generation, re-rendering, and fusion on DAVIS car-turn (\x8, $\tau=1/8$).} The top row shows full frames with PSNR$\uparrow$ / LPIPS$\downarrow$ computed over the entire image; the bottom enlarges the same cropped region for each output. Arrows highlight duplicated contours near the front tire and fine details around the license plate and headlight.}
  \label{fig:carturn_stages}
  \vspace{-0.3cm}
\end{figure*}

\noindent\textbf{Fusion with Re-rendered Observations.}
We next evaluate whether re-rendered observations improve reconstruction beyond fusing the draft with generated appearance (\Cref{tab:components}).
\textit{Fusion w/o readback} retains the same fusion architecture and generated appearance candidate, replacing both re-rendered inputs $\mathbf{I}_{\mathrm{rf}}^{\mathrm{warp}}$ and $\mathbf{I}_{g}^{\mathrm{warp}}$ with the draft $\mathbf{I}^{\mathrm{fs}}$. 
Full reconstruction improves PSNR over this control by  0.14, 0.27, and 0.20\,dB and reduces LPIPS by 0.0030, 0.0037, and 0.0042 at \x4, \x8, and \x16.
It also outperforms generation alone and re-rendering alone by 0.39--2.15\,dB and 0.96--1.46\,dB, respectively, with lower LPIPS.

Fig.~\ref{fig:carturn_stages} illustrates these complementary roles on DAVIS car-turn.
$\mathbf{I}^{\mathrm{gen}}$ reduces the draft's duplicated car contour but oversmooths license-plate and headlight details, whereas re-rendering retains observed details with residual ghosting near the front tire. The \textit{Full reconstruction} combines these cues, reducing visible ghosting while preserving clearer license-plate, headlight, and grille details than \textit{Fusion w/o readback}.

\noindent\textbf{Appearance Fusion Design.}
Keeping the generated and re-rendered inputs fixed, we examine how the fusion head combines them (\Cref{tab:btd}).
Free residual fusion predicts an unconstrained RGB correction to $\mathbf{I}_{\mathrm{rf}}^{\mathrm{warp}}$ without explicit mixing. The other controls replace the spatial mixing maps with calibrated constants or remove residual correction.
The full head achieves the highest PSNR and lowest LPIPS at all three factors, improving PSNR over free residual fusion by 0.12, 0.15, and 0.11\,dB at \x4, \x8, and \x16, respectively.
Fixing the mixing weights or removing the residual also worsens both metrics, although the \x16 PSNR difference without the residual is small (0.01\,dB).

\begin{table}[t]
\centering
\caption{\textbf{Appearance fusion design.}
Evaluation on 89 LAVIB parent videos, averaged over three paired training runs.
Within each run, variants share the generator, inputs, pretrained U-Net body, and adaptation budget.
The initial output is $\mathbf{I}_{\mathrm{rf}}^{\mathrm{warp}}$ for free residual fusion and calibrated mixing for the other variants.}
\vspace{-0.2cm}
\label{tab:btd}
\scriptsize
\setlength{\tabcolsep}{1.5pt}
\renewcommand{\arraystretch}{1.1}
\resizebox{\columnwidth}{!}{
\begin{tabular}{lcccccc}
\toprule
\multirow{2}{*}{\textbf{Variant}} & \multicolumn{2}{c}{\textbf{\x4}} & \multicolumn{2}{c}{\textbf{\x8}} & \multicolumn{2}{c}{\textbf{\x16}} \\
\cmidrule(lr){2-3}\cmidrule(lr){4-5}\cmidrule(lr){6-7}
 & PSNR$\uparrow$ & LPIPS$\downarrow$ & PSNR$\uparrow$ & LPIPS$\downarrow$ & PSNR$\uparrow$ & LPIPS$\downarrow$ \\
\midrule
Spatial mixing, no residual & 35.85 & 0.0267 & 32.80 & 0.0393 & 27.92 & 0.0695 \\
Fixed mixing maps + residual & 35.83 & 0.0262 & 32.77 & 0.0405 & 27.90 & 0.0752 \\
Free residual fusion & 35.86 & 0.0253 & 32.70 & 0.0401 & 27.82 & 0.0768 \\
\midrule
\oursrow
\textbf{Full reconstruction (Ours)} & \textbf{35.98} & \textbf{0.0240} & \textbf{32.85} & \textbf{0.0369} & \textbf{27.93} & \textbf{0.0676} \\
\bottomrule
\end{tabular}}
\end{table}

\begin{table}[!t]
\centering
\caption{\textbf{End-to-end inference cost.} Latency is total inference time per synthesized frame, including auxiliary processing and excluding model loading on a B200 GPU. NFE counts primary-network evaluations in the stated units. Memory is the maximum peak allocation across factors.}
\vspace{-0.2cm}
\label{tab:cost}
\scriptsize
\setlength{\tabcolsep}{4pt}
\renewcommand{\arraystretch}{1.1}
\resizebox{\columnwidth}{!}{
\begin{tabular}{llccccc}
\toprule
\multirow{2}{*}{\textbf{Method}} & \multirow{2}{*}{\textbf{NFE}} & \multicolumn{4}{c}{\textbf{s / frame}$\downarrow$} & \multirow{2}{*}{\makecell{\textbf{Peak}\\\textbf{Memory}\\(GiB)}} \\
\cmidrule(lr){3-6}
 & & \x2 & \x4 & \x8 & \x16 & \\
\midrule
LDMVFI~\cite{ldmvfi2024} & 200 / frame & 4.283 & 4.783 & 4.338 & 4.361 & 13.42 \\
CBBD~\cite{cbbd2024} & 50 / frame & 1.172 & 1.054 & 1.237 & 1.190 & 17.87 \\
MoMo~\cite{momo2025} & 8 / frame & 0.087 & 0.090 & 0.088 & 0.085 & 11.02 \\
EDEN~\cite{eden2025} & 2 / frame & 0.101 & 0.093 & 0.093 & 0.102 & 7.56 \\
TLB-VFI~\cite{tlbvfi2025} & 10 / frame & 0.441 & 0.445 & 0.438 & 0.431 & 34.24 \\
SPEED~\cite{speed2026} & 1 / frame & 0.013 & 0.019 & 0.020 & 0.018 & 6.38 \\
\midrule
LDF-VFI~\cite{ldfvfi2026} & 16 / chunk & 10.304 & 7.048 & 6.093 & 7.305 & 11.91 \\
\oursrow \textbf{\method} & 1 / window & 0.319 & 0.217 & 0.186 & 0.175 & 42.05 \\
\bottomrule
\end{tabular}}
\vspace{-0.3cm}
\end{table}

\subsection{Efficiency}
\label{sec:efficiency}
\method{} evaluates the video generator once per window, sharing this computation across the target frames. Under the 49-frame, 720p/B200 protocol (Table~\ref{tab:cost}), the complete pipeline requires 0.175--0.319\,s per synthesized frame, including flow estimation, rendering, and fusion, achieving a 32--42$\times$ speedup over the evaluated 16-step LDF-VFI~\cite{ldfvfi2026}. This reduction comes with higher peak allocated memory (42.05 vs.\ 11.91\,GiB), while MoMo, EDEN, and SPEED have lower latency.
At \x8, enabling readback adds 0.050\,s per synthesized frame, resulting in a total latency of 0.186\,s/frame. For this factor, Table~\ref{tab:components} shows that \textit{Full reconstruction (Ours)} improves PSNR by 0.27\,dB and reduces LPIPS by 0.0037 over \textit{Fusion w/o readback} on SNU-FILM.

\subsection{Limitations}
\label{sec:limitations}
Our reconstruction remains sensitive to the accuracy of the generated reference and the estimated correspondences, as errors in either can propagate to the re-rendered observations and persist after fusion. The framework also retains substantial computational and memory requirements: although one-step generation avoids repeated sampling, the video backbone and subsequent reconstruction stages remain costly compared with lightweight VFI methods.
\section{Conclusion}
\label{sec:conclusion}
We presented RePix-VFI, a multi-frame VFI framework that couples draft-conditioned one-step video diffusion with reconstruction from observed keyframes.
Generative readback matches observed keyframes to generated predictions using a frozen optical-flow estimator, providing correspondences for re-rendering their original RGB pixels.
A lightweight appearance fusion head combines these renderings with generated appearance.
Experiments on SNU-FILM and DAVIS demonstrate improved reconstruction, perceptual quality, and temporal consistency, while controlled ablations further support the use of generated correspondence references. This design allows generated predictions to guide the placement of observed pixels without requiring the generator to reproduce all available appearance details.

\clearpage

\clearpage
\appendix

\begin{table}[h]
\small
\centering
\caption{\textbf{Overview of the \textit{Appendix}.}}
\vspace{-0.2cm}
\label{tab:supple_overview}
\setlength{\tabcolsep}{6pt}
\renewcommand{\arraystretch}{1.1}
\resizebox{0.8\linewidth}{!}{%
\begin{tabular}{ll}
\toprule
\rowcolor{cyan!8}
\textbf{Section} & \textbf{Contents} \\
\midrule
Sec.~\ref{supp:impl} & Implementation and evaluation details \\
Sec.~\ref{supp:controlled} & Component analysis \\
Sec.~\ref{supp:results} & Benchmark results and efficiency \\
Sec.~\ref{supp:qual} & Visual results \\
Sec.~\ref{supp:limitations} & Limitations and future work \\
\bottomrule
\end{tabular}}
\end{table}

\noindent This supplement analyzes how one-step video predictions guide correspondence estimation for re-rendering original keyframe RGB outside the VAE.
It includes implementation details, component controls, benchmark and visual comparisons, efficiency measurements, and limitations (Table~\ref{tab:supple_overview}).

\section{Implementation and Evaluation Details}
\label{supp:impl}

\noindent\textbf{Draft and generator.}
\label{supp:generator}
We construct the flow-rendered draft using flows estimated by frozen RAFT-Large at half the output height and width. Readback flows are estimated at full output resolution.
The generator takes VAE encodings of the flow-rendered draft $\mathbf I^{\mathrm{fs}}$ and the observation sequence $\mathbf A$, concatenated with the observation mask $\mathbf m$ to form a 100-channel input ($48+48+4$).
Text conditioning uses a fixed empty-prompt embedding.
We adapt the generator using AdamW with a learning rate of $5\times10^{-5}$ and a batch size of 24, maintaining an EMA with decay 0.999.
During latent-stage adaptation, the frozen Wan model~\cite{wan2025} provides the reference score; an auxiliary model with separate LoRA adapters is trained with a flow-matching objective on detached generated latents.
A framewise 2D discriminator operating in latent space supplies the adversarial loss. Pixel-stage adaptation uses VGG-LPIPS~\cite{lpips2018}.

\noindent\textbf{Fusion training.}
\label{supp:fusion_train}
We train the 1.19M-parameter appearance fusion head using AdamW with a learning rate of $10^{-4}$.
Training consists of 10.5k RGB-residual pretraining updates followed by 3k adaptation updates.
For the pair in Table~\ref{tab:components} of the main paper, \textit{Full reconstruction} and \textit{Fusion w/o readback} share the frozen generator, training data, optimization settings, and adaptation budget.
Both are initialized from the same U-Net body, pretrained using the inputs in Eq.~\eqref{eq:fusion_head} of the main paper.
The output projection is zero-initialized, so adaptation starts from the mixing weights given by the fixed logit offsets and from a zero residual $\mathbf R$, which is bounded by $\tanh$.
\label{supp:fusion_loss}
During adaptation, the perceptual term in Eq.~\eqref{eq:fusion_loss} of the main paper combines relative VGG-LPIPS with $\lambda(d)$-weighted relative AlexNet-LPIPS.
Here, $d=\min_{k\in\mathcal K_s}|t-k|$ is the distance in frames to the nearest observed keyframe. The weight $\lambda(d)$ increases linearly from zero at $d=1$ to one at $d=4$ and remains one thereafter.

\noindent\textbf{Calibration.}
Calibration uses the same LAVIB~\cite{lavib2024} validation segments as fusion training.
The 48 parent videos in Table~\ref{tab:supp_draft_linear} and the 89 parent videos in Table~\ref{tab:btd} of the main paper are disjoint from each other and from the generator-adaptation, fusion-training, and calibration sets at the LAVIB parent-ID level.
Grid search selects $\alpha_s$ by minimizing relative MSE plus twice relative VGG-LPIPS, and the logit offsets by maximizing mean PSNR.
\textit{Fusion w/o readback} is calibrated separately.
Both $\alpha_s$ and the logit offsets remain fixed during adaptation and evaluation.

\noindent\textbf{Evaluation.}
\label{supp:inference}
RePix-VFI uses native-resolution 49-frame windows (stride 48), with shorter windows ending at observed keyframes for short clips.
For each factor, all methods are evaluated on the same target frames, excluding frames after the last observed keyframe.
We compute tOF~\cite{tecogan2020} using RAFT-Small~\cite{raft2020}. The tLP metric~\cite{tecogan2020} is the absolute difference between adjacent-frame AlexNet-LPIPS distances in the reconstructed and GT videos, averaged over all pairs and scaled by 100.
For FVD and temporal metrics on DAVIS, REDS, and Adobe240, observed-frame outputs are replaced with GT, while SNU-FILM tOF uses each method’s observed-frame outputs.
\section{Component Analysis}
\label{supp:controlled}

\subsection{Source Initialization}
\label{supp:draft_linear}
Table~\ref{tab:supp_draft_linear} compares a linear RGB blend of the bounding observations with the flow-rendered draft $\mathbf{I}^{\mathrm{fs}}$ as the generator's source initialization.
Conditioning and optimization settings are shared, following Sec.~\ref{supp:impl}.
At a fixed budget of 5k latent-stage updates, the flow-rendered source yields higher generated-frame PSNR (by 1.57--5.50\,dB) and lower LPIPS than linear RGB initialization at every factor.
This comparison uses 49-frame $512\times512$ windows and precedes pixel-stage adaptation, readback, and appearance fusion.
Separately, pixel-stage adaptation raises the PSNR of $\mathbf I^{\mathrm{gen}}$ by 1.44--1.84\,dB on SNU-FILM (Table~\ref{tab:components} of the main paper).

\begin{table}[!t]
\centering
\caption{\textbf{Linear versus flow-rendered draft initialization.}
Results on 48 held-out LAVIB parent videos. Generated rows use 5k latent-stage updates only (one run per source); output differences are draft minus linear.}
\label{tab:supp_draft_linear}
\small
\setlength{\tabcolsep}{5pt}
\renewcommand{\arraystretch}{1.1}
\resizebox{1.0\linewidth}{!}{%
\begin{tabular}{lccccc}
\toprule
\textbf{Output} & \textbf{Metric} & \textbf{\x2} & \textbf{\x4} & \textbf{\x8} & \textbf{\x16} \\
\midrule
\multirow{2}{*}{\makecell[l]{Linear source\\(input)}}
 & PSNR$\uparrow$ & 28.19 & 25.12 & 22.64 & 20.45 \\
 & LPIPS$\downarrow$ & 0.1058 & 0.1161 & 0.1550 & 0.2192 \\
\addlinespace[2pt]
\multirow{2}{*}{\makecell[l]{Generated from\\linear source}}
 & PSNR$\uparrow$ & 30.31 & 28.45 & 27.82 & 24.79 \\
 & LPIPS$\downarrow$ & 0.0674 & 0.0769 & 0.0871 & 0.1255 \\
\addlinespace[2pt]
\multirow{2}{*}{\makecell[l]{Flow-rendered draft $\mathbf{I}^{\mathrm{fs}}$ \\ (input)}}
 & PSNR$\uparrow$ & 37.61 & 35.31 & 32.11 & 26.68 \\
 & LPIPS$\downarrow$ & 0.0262 & 0.0350 & 0.0554 & 0.1116 \\
\addlinespace[2pt]
\multirow{2}{*}{\makecell[l]{Generated from $\mathbf{I}^{\mathrm{fs}}$}}
 & PSNR$\uparrow$ & 35.60 & 33.95 & 31.35 & 26.36 \\
 & LPIPS$\downarrow$ & 0.0258 & 0.0351 & 0.0561 & 0.1122 \\
\midrule
\oursrow
 & {${\Delta}$PSNR} & \textbf{+5.29} & \textbf{+5.50} & \textbf{+3.53} & \textbf{+1.57} \\
\oursrow
\multirow{-2}{*}{\makecell[l]{Generated-output\\difference}}
 & {${\Delta}$LPIPS} & \textbf{-0.0416} & \textbf{-0.0418} & \textbf{-0.0310} & \textbf{-0.0133} \\
\bottomrule
\end{tabular}}
\end{table}

\subsection{Correspondence Reference and Keyframe RGB Source}
\label{supp:readback_controls}
The controls in Table~\ref{tab:supp_public_reference} support using generated predictions for correspondence estimation and original observations for re-rendering.
Table~\ref{tab:supp_public_reference}(a) summarizes the reference control in Table~\ref{tab:components} of the main paper.
Table~\ref{tab:supp_public_reference}(b) replaces original keyframe RGB $I_k$ with $\mathcal{D}(\mathcal{E}(I_k))$, applying the frozen VAE to each keyframe as a one-frame video.
With flows and the renderer fixed, using original keyframe RGB rather than its VAE reconstruction improves pre-fusion PSNR by 0.50–1.98 dB and lowers LPIPS at every factor.

\begin{table}[!t]
\small
\centering
\caption{\textbf{Correspondence-reference and RGB-source controls on SNU-FILM.}
Pre-fusion quality differences in $\mathbf I_{\mathrm{rf}}^{\mathrm{warp}}$. (a) fixes the flow estimator and rendering settings; (b) additionally fixes the estimated flows. Positive $\Delta$PSNR and negative $\Delta$LPIPS favor the first named choice.}
\label{tab:supp_public_reference}
\label{tab:supp_vae_observations}
\setlength{\tabcolsep}{7pt}
\renewcommand{\arraystretch}{1.1}
\resizebox{0.8\linewidth}{!}{%
\begin{tabular}{lccc}
\toprule
\textbf{Metric} & \textbf{\x4} & \textbf{\x8} & \textbf{\x16} \\
\midrule
\rowcolor{gray!8}
\multicolumn{4}{l}{\textbf{(a) Generated $-$ draft correspondence reference}} \\
$\Delta$PSNR (dB)$\uparrow$ & +0.45 & +0.72 & +0.57 \\
$\Delta$LPIPS$\downarrow$ & -0.0009 & -0.0019 & -0.0031 \\
\midrule
\rowcolor{gray!8}
\multicolumn{4}{l}{\textbf{(b) Original $-$ VAE-reconstructed keyframe RGB}} \\
$\Delta$PSNR (dB)$\uparrow$ & +1.98 & +1.08 & +0.50 \\
$\Delta$LPIPS$\downarrow$ & -0.0190 & -0.0172 & -0.0150 \\
\bottomrule
\end{tabular}}
\end{table}

\subsection{Fusion Training and Reference Choice}
\label{supp:training_resolution}\label{supp:reported_reconstruction}
Table~\ref{tab:supp_public_history}(a) tests two fusion-training configurations with three paired runs each.
One pretrains each U-Net body with its own inputs; the other shares no-readback pretraining and uses full-resolution draft flows for fusion.
Both retain the generator's input draft. Each pair shares the generator, training data, objective, and adaptation budget.
Readback lowers mean LPIPS at every factor; PSNR improves at \x4 and \x8 and changes by less than 0.04\,dB at \x16.
The reference control in Table~\ref{tab:supp_public_reference}(a) is measured before appearance fusion and does not depend on these configurations.

\label{supp:gimm_reference}
Table~\ref{tab:supp_public_history}(b) uses GIMM-VFI predictions as correspondence references while retaining $\mathbf I^{\mathrm{gen}}$ for appearance fusion. Rendering settings, fusion initialization, the adaptation budget, and the calibration procedure are shared (Sec.~\ref{supp:impl}).

GIMM-VFI uses native two-frame inputs and has higher RGB PSNR than $\mathbf I^{\mathrm{gen}}$ at \x4 and \x8 (Tables~\ref{tab:snufilm} and~\ref{tab:components} of the main paper).
Yet generated references yield higher final PSNR at these factors and lower LPIPS at all factors. Thus, RGB PSNR alone does not determine a prediction’s usefulness as a correspondence reference.

\begin{table}[!t]
\small
\centering
\caption{\textbf{Readback and reference controls on SNU-FILM.}
Mean paired differences after appearance fusion over three training runs.}
\label{tab:supp_public_history}
\label{tab:supp_joint_history}
\label{tab:supp_reference_alternative}
\setlength{\tabcolsep}{7pt}
\renewcommand{\arraystretch}{1.1}
\resizebox{0.8\linewidth}{!}{%
\begin{tabular}{lccc}
\toprule
\textbf{Metric} & \textbf{\x4} & \textbf{\x8} & \textbf{\x16} \\
\midrule
\rowcolor{gray!8}
\multicolumn{4}{l}{\textbf{(a) Readback $-$ no readback}} \\
\multicolumn{4}{l}{Separate pretraining; half-res. draft flows} \\
$\Delta$PSNR (dB)$\uparrow$ & +0.033 & +0.160 & +0.007 \\
$\Delta$LPIPS$\downarrow$ & -0.0009 & -0.0023 & -0.0026 \\
\addlinespace[4pt]
\multicolumn{4}{l}{Shared no-readback pretraining; full-res. draft flows} \\
$\Delta$PSNR (dB)$\uparrow$ & +0.009 & +0.077 & -0.037 \\
$\Delta$LPIPS$\downarrow$ & -0.0006 & -0.0024 & -0.0026 \\
\midrule
\rowcolor{gray!8}
\multicolumn{4}{l}{\textbf{(b) Generated $-$ GIMM-VFI reference}} \\
$\Delta$PSNR (dB)$\uparrow$ & +0.089 & +0.153 & -0.005 \\
$\Delta$LPIPS$\downarrow$ & -0.0004 & -0.0019 & -0.0030 \\
\bottomrule
\end{tabular}}
\end{table}

\subsection{Fusion Design on Public Benchmarks}
\label{supp:fusion_public}
In Table~\ref{tab:supp_fusion_public}, the full appearance fusion head achieves the lowest mean LPIPS at all SNU-FILM and Adobe240 settings, and the highest mean PSNR at \x4 and \x8 on both datasets.

\begin{table}[!t]
\centering\small
\caption{\textbf{Fusion design on public benchmarks.}
Three-run means on (a) SNU-FILM (31 clips) and (b) Adobe240 (17 sequences), using the heads from Table~4 of the main paper without further adaptation. \textbf{Bold} and \underline{underline} indicate the best and second-best results.}
\label{tab:supp_fusion_public}
\begingroup
\scriptsize
\setlength{\tabcolsep}{1.5pt}
\renewcommand{\arraystretch}{1.1}
\resizebox{\columnwidth}{!}{%
\begin{tabular}{lcccccc}
\toprule
\multirow{2}{*}{\textbf{Variant}} & \multicolumn{2}{c}{\textbf{\x4}} & \multicolumn{2}{c}{\textbf{\x8}} & \multicolumn{2}{c}{\textbf{\x16}} \\
\cmidrule(lr){2-3}\cmidrule(lr){4-5}\cmidrule(lr){6-7}
 & PSNR$\uparrow$ & LPIPS$\downarrow$ & PSNR$\uparrow$ & LPIPS$\downarrow$ & PSNR$\uparrow$ & LPIPS$\downarrow$ \\
\midrule
\rowcolor{gray!8}
\multicolumn{7}{l}{\textbf{(a) SNU-FILM}} \\
Spatial mixing, no residual & \underline{36.81} & 0.0249 & \underline{33.27} & 0.0368 & \textbf{28.44} & 0.0642 \\
Fixed mixing maps + residual & 36.67 & 0.0220 & 33.16 & 0.0349 & 28.37 & 0.0639 \\
Free residual fusion & 36.75 & \underline{0.0211} & 33.11 & \underline{0.0340} & 28.31 & \underline{0.0636} \\
\oursrow
\textbf{Full reconstruction (Ours)} & \textbf{36.95} & \textbf{0.0203} & \textbf{33.30} & \textbf{0.0325} & \underline{28.42} & \textbf{0.0607} \\
\midrule
\rowcolor{gray!8}
\multicolumn{7}{l}{\textbf{(b) Adobe240}} \\
Spatial mixing, no residual & 35.21 & 0.0280 & 33.13 & 0.0339 & \textbf{27.46} & 0.0585 \\
Fixed mixing maps + residual & 35.31 & \underline{0.0235} & 33.16 & 0.0301 & 27.43 & 0.0558 \\
Free residual fusion & \underline{35.36} & \textbf{0.0230} & \underline{33.18} & \underline{0.0294} & 27.41 & \underline{0.0550} \\
\oursrow
\textbf{Full reconstruction (Ours)} & \textbf{35.39} & \textbf{0.0230} & \textbf{33.22} & \textbf{0.0292} & \underline{27.44} & \textbf{0.0544} \\
\bottomrule
\end{tabular}}
\endgroup
\end{table}

\section{Benchmark Results and Efficiency}
\label{supp:results}\label{supp:robustness_cost}

\subsection{Extended Metrics on SNU-FILM and DAVIS}
\label{supp:temporal_controls}
Table~\ref{tab:supp_snufilm_fvd} adds FVD~\cite{fvd2018} to the SNU-FILM~\cite{cain2020} comparison, and Table~\ref{tab:davis_full} extends the DAVIS~\cite{davis2016} comparison to all 15 baselines and seven metrics.
On DAVIS, RePix-VFI leads in PSNR, SSIM, LPIPS, FloLPIPS, tOF, and tLP at all three factors, while ranking second or third in FVD.
On SNU-FILM, it ranks second in FVD at \x8 and \x16.

\begin{table}[!t]
\centering
\small
\caption{\textbf{Video-level FVD comparison on SNU-FILM.}
Lower FVD is better. \textbf{Bold} and \underline{underline} indicate the best and second-best results.}
\label{tab:supp_snufilm_fvd}
\setlength{\tabcolsep}{7pt}
\renewcommand{\arraystretch}{1.1}
\resizebox{0.65\columnwidth}{!}{%
\begin{tabular}{lccc}
\toprule
\textbf{Method} & \textbf{\x4} & \textbf{\x8} & \textbf{\x16} \\
\midrule
\grouprow{4}{Non-diffusion}
FILM~\cite{film2022} & \textbf{4.17} & 15.90 & 36.90 \\
RIFE~\cite{rife2022} & 5.95 & 18.68 & 37.35 \\
AMT~\cite{amt2023} & 6.63 & 19.39 & 60.22 \\
GIMM-VFI~\cite{gimmvfi2024} & 7.20 & 22.10 & 70.67 \\
InterpAny~\cite{interpany2024} & 9.24 & 22.28 & 55.59 \\
BiM-VFI~\cite{bimvfi2025} & 5.68 & 17.20 & 35.79 \\
VTinker~\cite{vtinker2026} & 5.64 & 18.23 & 42.60 \\
EMA-VFI~\cite{emavfi2023} & 5.73 & 20.12 & 53.12 \\
\midrule
\grouprow{4}{Diffusion-based}
LDMVFI~\cite{ldmvfi2024} & 8.29 & 26.81 & 66.89 \\
CBBD~\cite{cbbd2024} & 5.09 & 19.56 & 55.22 \\
MoMo~\cite{momo2025} & 4.75 & 19.68 & 44.68 \\
EDEN~\cite{eden2025} & 9.49 & 25.20 & 59.35 \\
TLB-VFI~\cite{tlbvfi2025} & \underline{4.42} & 17.91 & 38.60 \\
SPEED~\cite{speed2026} & 6.53 & 22.97 & 51.91 \\
LDF-VFI~\cite{ldfvfi2026} & 12.50 & \textbf{14.98} & \textbf{25.89} \\
\oursrow
\textbf{\method} & 5.35 & \underline{15.42} & \underline{35.53} \\
\bottomrule
\end{tabular}}
\end{table}

\noindent\textbf{Observed-frame control.}
The SNU-FILM tOF advantage over LDF-VFI remains when observed frames are replaced with GT: LDF-VFI obtains $0.3177$, $0.4118$, and $0.5917$ at \x4, \x8, and \x16, respectively, versus RePix-VFI's $0.2474$, $0.3490$, and $0.5400$.

\subsection{Generalization to REDS and Adobe240}
We evaluate on the REDS validation set~\cite{reds2019} and the Adobe240~\cite{adobe2402017} test split from VideoINR~\cite{videoinr2022}, without retraining or recalibration (Tables~\ref{tab:supp_reds} and~\ref{tab:supp_adobe}).
RePix-VFI achieves the highest PSNR and lowest tOF on both datasets at all three factors.
It also obtains the lowest FloLPIPS on REDS and the best results in all five metrics on Adobe240 at \x8 and \x16.
Where it does not achieve the lowest LPIPS, its score is within 0.001 of the best result.

\begin{table}[!t]
\centering\small
\caption{\textbf{Reconstruction stage costs.}
Separately synchronized profiling in seconds per 49-frame, 720p window at \x8 on an NVIDIA B200. The median is computed separately for each stage and for the total runtime.}
\label{tab:supp_stage_cost}
\setlength{\tabcolsep}{7pt}
\renewcommand{\arraystretch}{1.1}
\resizebox{0.7\linewidth}{!}{%
\begin{tabular}{lc}
\toprule
\textbf{Stage} & \textbf{Time (s)}\\
\midrule
Draft construction & 0.71\\
One-step generation and VAE & 4.40\\
Readback and re-rendering & 2.21\\
Fusion and remaining processing & 0.57\\
\midrule
Total (instrumented run) & 7.90\\
\bottomrule
\end{tabular}}
\end{table}

\subsection{Runtime and Memory}
\label{supp:extended_runtime}
We follow the 49-frame, 720p/B200 protocol in Table~5 of the main paper: median end-to-end latency after warm-up, excluding model loading.
Per-frame latency is computed by dividing window time by the number of synthesized frames: $24$, $36$, $42$, and $45$ at \x2, \x4, \x8, and \x16, respectively.
Methods retain their native numerical precision.
LDF-VFI uses 16 sampling steps and fixed 40-frame increments in the 49-frame wrapper; quality evaluation uses its native entire-video script.
For RePix-VFI, the median end-to-end latency at \x8 is 7.81 \,s/window (0.186\,s/synthesized frame).
A separate profiling run with synchronization between stages gives a median total of 7.90 s (Table~\ref{tab:supp_stage_cost}). Generation and VAE processing account for approximately 56\% of the total, and readback and re-rendering for 28\%.
Peak memory is reported in Table~\ref{tab:cost} of the main paper and measures allocated CUDA memory, excluding unused reservations.

\section{Visual Results}
\label{supp:qual}

\begin{figure*}[!t]
\centering
\includegraphics[width=\textwidth]{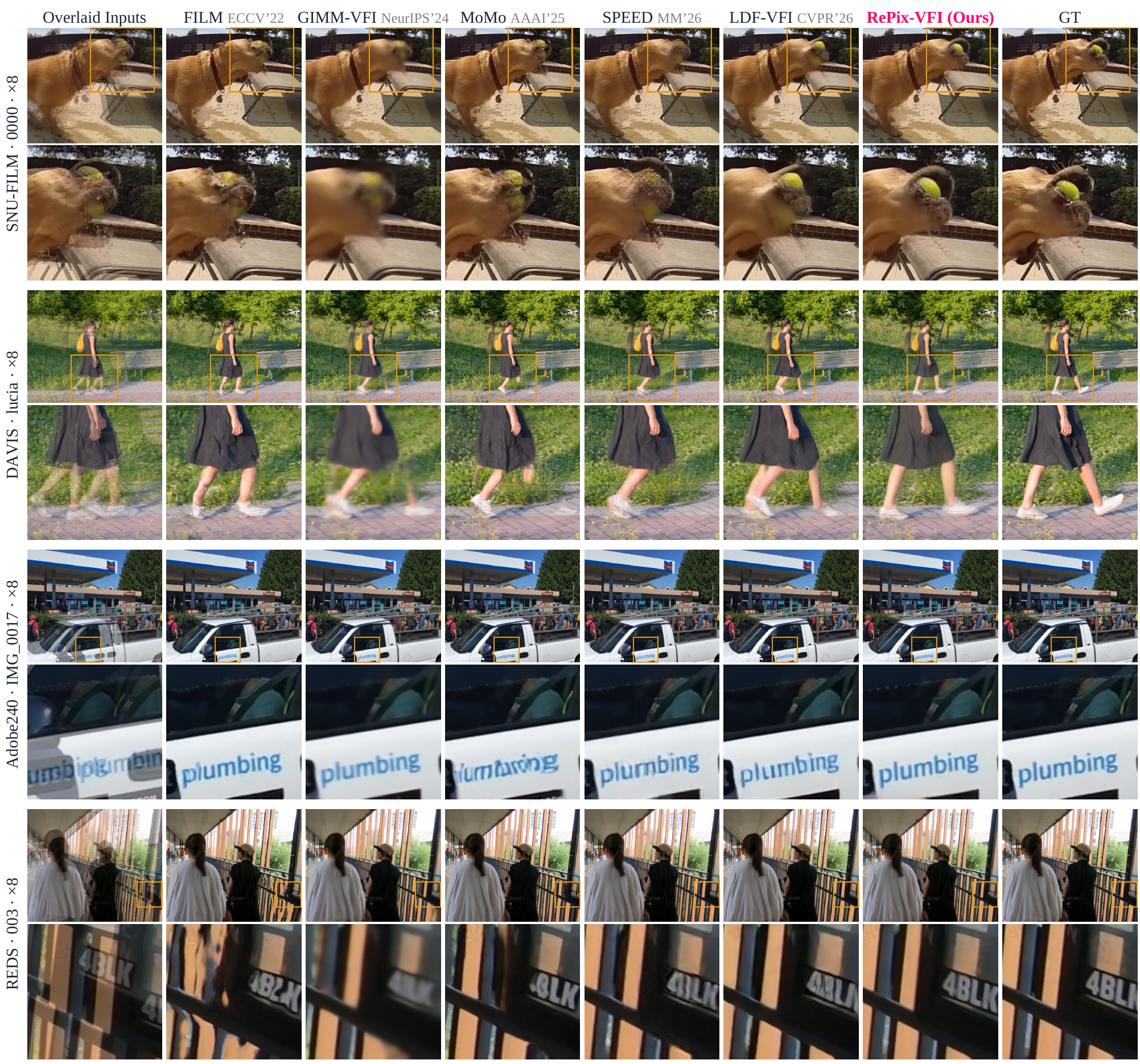}
\caption{\textbf{Qualitative comparisons at \x8.}
Midpoint targets ($\tau=0.5$) from SNU-FILM, DAVIS, Adobe240, and REDS, from top to bottom. The first column shows an equal-weight overlay of the bounding observations; subsequent columns show predictions and GT. Boxes indicate the regions enlarged below.}
\label{fig:supp_datasets_x8}
\end{figure*}

\begin{figure*}[!t]
\centering
\captionsetup{hypcap=false}
\includegraphics[width=\textwidth]{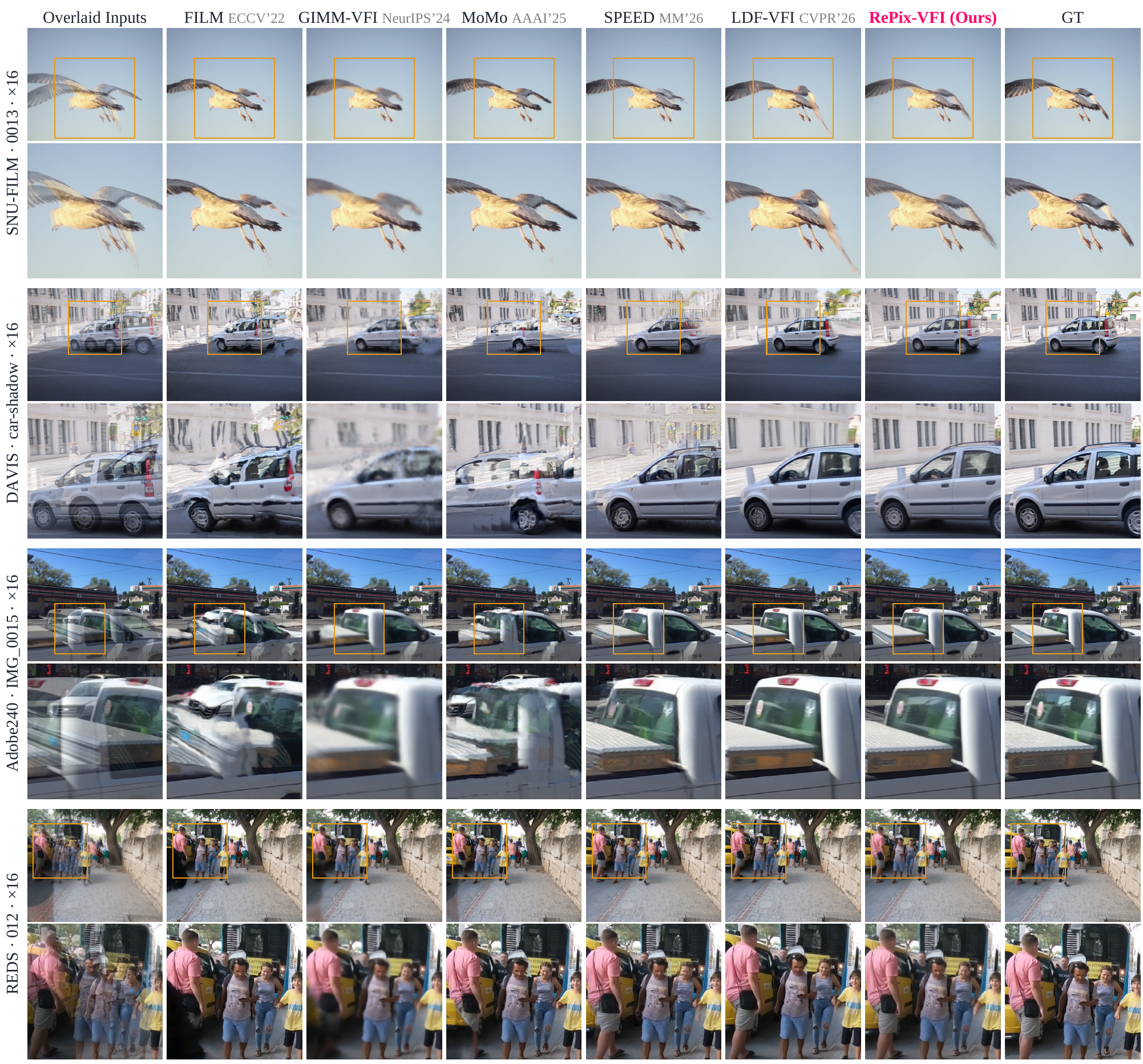}
\captionof{figure}{\textbf{Qualitative comparisons at \x16.}
Midpoint targets ($\tau=0.5$) from SNU-FILM, DAVIS, Adobe240, and REDS, from top to bottom. The first column shows an equal-weight overlay of the bounding observations; subsequent columns show predictions and GT. Boxes indicate the regions enlarged below.}
\label{fig:supp_datasets_x16}
\end{figure*}

Fig.~\ref{fig:supp_qualitative_full} extends the DAVIS motorbike example in Fig.~\ref{fig:qualitative} of the main paper across target times.
RePix-VFI closely follows the changing object positions in GT as the motorcycle passes behind the planter.
At \x8, RePix-VFI preserves distinct ball and mouth contours on SNU-FILM, leg separation on DAVIS, and legible lettering on Adobe240 and REDS (Fig.~\ref{fig:supp_datasets_x8}).
At \x16, it preserves wing shape, vehicle contours, and separation between nearby pedestrians (Fig.~\ref{fig:supp_datasets_x16}).

Accompanying demo videos show all qualitative examples from the main paper and this supplement at playback rates of 24, 8, and 4 fps.

\section{Limitations and future work}
\label{supp:limitations}
\begin{figure*}[!t]
\centering
\includegraphics[width=\textwidth]{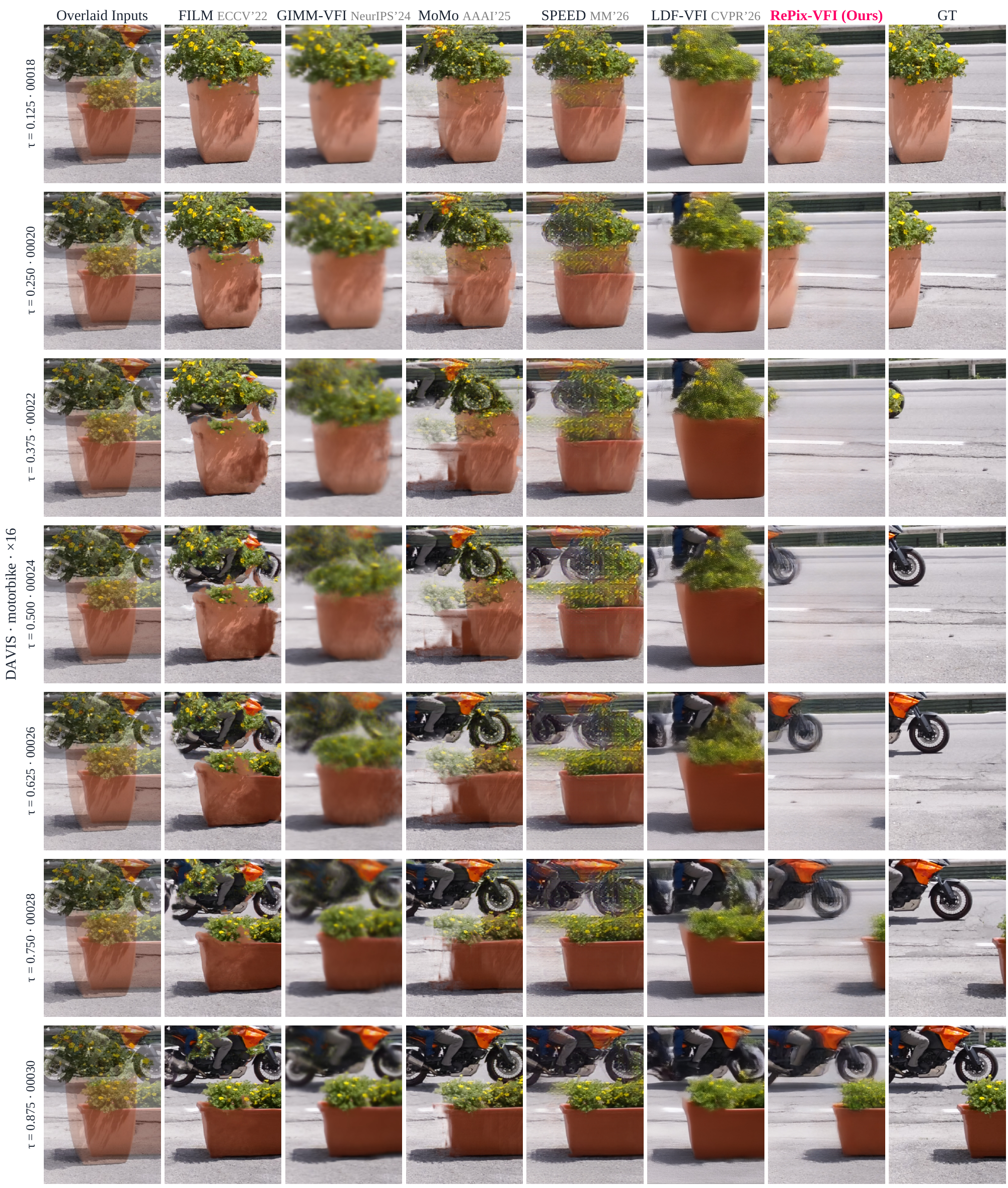}
\caption{\textbf{Multiple target times on DAVIS motorbike at \x16.}
Targets 00018--00030 in steps of two lie between observed keyframes 00016 and 00032 ($\tau=0.125,0.250,\ldots,0.875$).
The first column shows an equal-weight overlay of the bounding observations. All methods and target times use the same fixed crop.}
\label{fig:supp_qualitative_full}
\end{figure*}

Reconstructing fine structures under fast motion remains challenging, as illustrated by the wheel blur at $\tau=0.750$ in Fig.~\ref{fig:supp_qualitative_full}.
In the accompanying DAVIS car-roundabout and car-shadow demo videos, wheels can show less rotation than in GT while the vehicle moves.
A temporal loss focused on moving object parts could help recover such local motion.
A smaller video backbone and reduced-resolution readback could reduce the dominant costs in Table~\ref{tab:supp_stage_cost}.

\FloatBarrier

\clearpage
\begin{table*}[!p]
\centering
\caption{\textbf{Extended quantitative comparison on DAVIS 2016.}
Evaluation on all 50 sequences at native resolutions. Temporal metrics and FVD use GT observed frames (Sec.~\ref{supp:impl}). \textbf{Bold} and \underline{underline} indicate the best and second-best results.}
\vspace{-0.2cm}
\label{tab:davis_full}
\small
\setlength{\tabcolsep}{20pt}
\renewcommand{\arraystretch}{1.05}
\resizebox{\linewidth}{!}{%
\begin{tabular}{lccccccc}
\toprule
\textbf{Method} / \textbf{\x4} & PSNR$\uparrow$ & SSIM$\uparrow$ & LPIPS$\downarrow$ & FloLPIPS$\downarrow$ & tOF$\downarrow$ & tLP$\downarrow$ & FVD$\downarrow$ \\
\midrule
\grouprow{8}{Non-diffusion}
FILM~\cite{film2022} & 21.80 & 0.7587 & 0.2048 & 0.2509 & 23.79 & 6.286 & 169.41 \\
RIFE~\cite{rife2022} & 22.83 & 0.7882 & 0.1683 & 0.2171 & 17.80 & 4.291 & 118.50 \\
AMT~\cite{amt2023} & 21.76 & 0.7682 & 0.2999 & 0.3687 & 28.20 & 6.546 & 237.96 \\
GIMM-VFI~\cite{gimmvfi2024} & \underline{23.81} & \underline{0.8209} & 0.2737 & 0.3293 & 15.67 & 6.322 & 145.37 \\
InterpAny~\cite{interpany2024} & 22.27 & 0.7812 & 0.3049 & 0.3785 & 23.77 & 6.349 & 242.35 \\
BiM-VFI~\cite{bimvfi2025} & 22.86 & 0.7910 & 0.1643 & 0.2130 & \underline{12.67} & \underline{3.059} & \textbf{81.34} \\
VTinker~\cite{vtinker2026} & 22.71 & 0.7818 & 0.1742 & 0.2218 & 18.68 & 4.410 & 112.24 \\
EMA-VFI~\cite{emavfi2023} & 22.61 & 0.7876 & 0.2743 & 0.3332 & 23.07 & 6.324 & 163.35 \\
\grouprow{8}{Diffusion-based}
LDMVFI~\cite{ldmvfi2024} & 20.87 & 0.7362 & 0.2499 & 0.3001 & 26.24 & 9.270 & 356.79 \\
CBBD~\cite{cbbd2024} & 20.60 & 0.7339 & 0.2634 & 0.3118 & 27.50 & 7.869 & 339.36 \\
MoMo~\cite{momo2025} & 22.50 & 0.7780 & 0.1718 & 0.2231 & 15.80 & 3.195 & 104.36 \\
EDEN~\cite{eden2025} & 22.34 & 0.7635 & 0.3260 & 0.3717 & 25.72 & 9.031 & 369.95 \\
TLB-VFI~\cite{tlbvfi2025} & 20.62 & 0.7322 & 0.2367 & 0.2838 & 27.53 & 10.192 & 306.43 \\
SPEED~\cite{speed2026} & 23.15 & 0.7951 & \underline{0.1570} & \underline{0.2022} & 12.74 & 4.219 & 114.93 \\
LDF-VFI~\cite{ldfvfi2026} & 22.74 & 0.7894 & 0.2444 & 0.2885 & 18.82 & 5.668 & 103.08 \\
\oursrow
\textbf{RePix-VFI} & \textbf{24.56} & \textbf{0.8263} & \textbf{0.1422} & \textbf{0.1785} & \textbf{8.11} & \textbf{2.672} & \underline{90.44} \\
\midrule
\textbf{Method} / \textbf{\x8} & PSNR$\uparrow$ & SSIM$\uparrow$ & LPIPS$\downarrow$ & FloLPIPS$\downarrow$ & tOF$\downarrow$ & tLP$\downarrow$ & FVD$\downarrow$ \\
\midrule
\grouprow{8}{Non-diffusion}
FILM~\cite{film2022} & 18.87 & 0.6755 & 0.2918 & 0.3348 & 28.38 & 12.758 & 402.86 \\
RIFE~\cite{rife2022} & 19.71 & 0.7035 & 0.2553 & 0.3015 & 24.74 & 9.505 & 274.82 \\
AMT~\cite{amt2023} & 19.12 & 0.6974 & 0.4137 & 0.4706 & 31.22 & 10.274 & 463.73 \\
GIMM-VFI~\cite{gimmvfi2024} & \underline{20.59} & \underline{0.7437} & 0.3903 & 0.4424 & 22.94 & 10.607 & 299.87 \\
InterpAny~\cite{interpany2024} & 19.53 & 0.7081 & 0.4108 & 0.4680 & 28.00 & 9.134 & 432.54 \\
BiM-VFI~\cite{bimvfi2025} & 19.71 & 0.7048 & 0.2451 & 0.2933 & \underline{20.65} & \underline{5.083} & \underline{186.78} \\
VTinker~\cite{vtinker2026} & 19.47 & 0.6910 & 0.2680 & 0.3160 & 25.83 & 10.453 & 304.52 \\
EMA-VFI~\cite{emavfi2023} & 19.76 & 0.7146 & 0.3933 & 0.4470 & 27.53 & 9.802 & 373.10 \\
\grouprow{8}{Diffusion-based}
LDMVFI~\cite{ldmvfi2024} & 18.67 & 0.6675 & 0.3200 & 0.3612 & 29.49 & 16.519 & 524.84 \\
CBBD~\cite{cbbd2024} & 18.29 & 0.6601 & 0.3492 & 0.3899 & 29.73 & 12.802 & 553.53 \\
MoMo~\cite{momo2025} & 19.33 & 0.6888 & 0.2638 & 0.3145 & 24.04 & 6.053 & 278.77 \\
EDEN~\cite{eden2025} & 20.06 & 0.6998 & 0.4239 & 0.4605 & 32.12 & 12.981 & 705.41 \\
TLB-VFI~\cite{tlbvfi2025} & 18.30 & 0.6591 & 0.3097 & 0.3508 & 29.72 & 16.402 & 506.87 \\
SPEED~\cite{speed2026} & 20.12 & 0.7126 & \underline{0.2426} & \underline{0.2881} & 21.31 & 8.222 & 283.50 \\
LDF-VFI~\cite{ldfvfi2026} & 20.11 & 0.7190 & 0.3055 & 0.3535 & 24.68 & 7.539 & \textbf{182.83} \\
\oursrow
\textbf{RePix-VFI} & \textbf{21.64} & \textbf{0.7549} & \textbf{0.2099} & \textbf{0.2455} & \textbf{13.53} & \textbf{4.784} & 192.52 \\
\midrule
\textbf{Method} / \textbf{\x16} & PSNR$\uparrow$ & SSIM$\uparrow$ & LPIPS$\downarrow$ & FloLPIPS$\downarrow$ & tOF$\downarrow$ & tLP$\downarrow$ & FVD$\downarrow$ \\
\midrule
\grouprow{8}{Non-diffusion}
FILM~\cite{film2022} & 17.06 & 0.6136 & 0.3692 & 0.4039 & 29.72 & 18.118 & 584.19 \\
RIFE~\cite{rife2022} & 17.54 & 0.6331 & 0.3475 & 0.3843 & 28.39 & 15.606 & 468.43 \\
AMT~\cite{amt2023} & 17.53 & 0.6471 & 0.5107 & 0.5515 & 31.21 & 15.768 & 793.09 \\
GIMM-VFI~\cite{gimmvfi2024} & 18.22 & \underline{0.6752} & 0.5046 & 0.5437 & 27.51 & 17.109 & 559.78 \\
InterpAny~\cite{interpany2024} & 17.81 & 0.6544 & 0.5028 & 0.5439 & 29.57 & 13.458 & 651.71 \\
BiM-VFI~\cite{bimvfi2025} & 17.40 & 0.6280 & 0.3413 & 0.3810 & \underline{26.06} & 9.471 & \underline{364.75} \\
VTinker~\cite{vtinker2026} & 17.39 & 0.6221 & 0.3575 & 0.3949 & 28.98 & 16.870 & 516.47 \\
EMA-VFI~\cite{emavfi2023} & 17.99 & 0.6606 & 0.4990 & 0.5417 & 29.22 & 15.252 & 662.71 \\
\grouprow{8}{Diffusion-based}
LDMVFI~\cite{ldmvfi2024} & 17.23 & 0.6159 & 0.3813 & 0.4115 & 30.08 & 21.370 & 711.21 \\
CBBD~\cite{cbbd2024} & 16.77 & 0.6051 & 0.4291 & 0.4591 & 30.10 & 15.504 & 756.59 \\
MoMo~\cite{momo2025} & 17.26 & 0.6189 & 0.3523 & 0.3936 & 28.33 & 10.935 & 509.24 \\
EDEN~\cite{eden2025} & \underline{18.38} & 0.6488 & 0.4940 & 0.5217 & 32.87 & 17.676 & 1355.58 \\
TLB-VFI~\cite{tlbvfi2025} & 17.14 & 0.6195 & 0.3630 & 0.3984 & 29.40 & 17.767 & 543.80 \\
SPEED~\cite{speed2026} & 17.85 & 0.6413 & \underline{0.3399} & \underline{0.3764} & 27.30 & 13.932 & 547.51 \\
LDF-VFI~\cite{ldfvfi2026} & 17.71 & 0.6480 & 0.3974 & 0.4405 & 26.97 & \underline{9.376} & \textbf{348.81} \\
\oursrow
\textbf{RePix-VFI} & \textbf{19.17} & \textbf{0.6848} & \textbf{0.2962} & \textbf{0.3275} & \textbf{19.92} & \textbf{8.464} & 391.26 \\
\bottomrule
\end{tabular}}
\end{table*}

\begin{table*}[!t]
\centering
\caption{\textbf{Quantitative comparison on REDS.}
Evaluation on all 30 official 120-fps validation sequences at native
$1280\times720$. \textbf{Bold} and \underline{underline} indicate the best and second-best results.}
\label{tab:supp_reds}
\setlength{\tabcolsep}{2.5pt}
\renewcommand{\arraystretch}{1.1}
\resizebox{\textwidth}{!}{%
\begin{tabular}{lccccccccccccccc}
\toprule
\multirow{2}{*}{\textbf{Method}}
 & \multicolumn{5}{c}{\textbf{REDS $\times4$}}
 & \multicolumn{5}{c}{\textbf{REDS $\times8$}}
 & \multicolumn{5}{c}{\textbf{REDS $\times16$}} \\
\cmidrule(lr){2-6}\cmidrule(lr){7-11}\cmidrule(lr){12-16}
 & PSNR$\uparrow$ & SSIM$\uparrow$ & LPIPS$\downarrow$
 & \textls[-45]{FloLPIPS}$\downarrow$ & tOF$\downarrow$
 & PSNR$\uparrow$ & SSIM$\uparrow$ & LPIPS$\downarrow$
 & \textls[-45]{FloLPIPS}$\downarrow$ & tOF$\downarrow$
 & PSNR$\uparrow$ & SSIM$\uparrow$ & LPIPS$\downarrow$
 & \textls[-45]{FloLPIPS}$\downarrow$ & tOF$\downarrow$ \\
\midrule
\grouprow{16}{Non-diffusion}
FILM~\cite{film2022}
 & 29.90 & 0.8963 & 0.0369 & 0.0397 & 0.5516
 & 24.21 & 0.7542 & 0.0752 & 0.0822 & 1.0115
 & 20.69 & 0.6471 & 0.1382 & 0.1573 & 1.5903 \\
RIFE~\cite{rife2022}
 & 29.09 & 0.8891 & 0.0410 & 0.0437 & 0.5907
 & 24.11 & 0.7544 & 0.0755 & 0.0808 & 1.0221
 & 20.69 & 0.6477 & 0.1342 & 0.1473 & 1.6355 \\
AMT~\cite{amt2023}
 & 30.05 & 0.8999 & 0.0456 & 0.0512 & 0.5873
 & 24.33 & 0.7604 & 0.0903 & 0.1070 & 1.1133
 & 20.81 & 0.6576 & 0.1835 & 0.2279 & 1.9164 \\
GIMM-VFI~\cite{gimmvfi2024}
 & \underline{30.59} & \underline{0.9086} & 0.0576 & 0.0618 & 0.5072
 & \underline{25.00} & \underline{0.7820} & 0.1224 & 0.1297 & 0.9322
 & \underline{21.55} & \textbf{0.6866} & 0.2332 & 0.2487 & 1.4521 \\
InterpAny~\cite{interpany2024}
 & 29.82 & 0.8979 & 0.0514 & 0.0573 & 0.5669
 & 24.37 & 0.7639 & 0.1046 & 0.1183 & 1.0552
 & 20.99 & 0.6653 & 0.2036 & 0.2343 & 1.6916 \\
BiM-VFI~\cite{bimvfi2025}
 & 29.79 & 0.8943 & 0.0396 & 0.0432 & 0.5878
 & 24.19 & 0.7529 & 0.0768 & 0.0831 & 1.0253
 & 20.77 & 0.6484 & 0.1366 & 0.1500 & 1.5256 \\
VTinker~\cite{vtinker2026}
 & 29.81 & 0.8954 & 0.0368 & \underline{0.0391} & 0.5459
 & 24.27 & 0.7546 & \textbf{0.0735} & \underline{0.0786} & 0.9834
 & 20.87 & 0.6501 & \underline{0.1328} & \underline{0.1451} & 1.5073 \\
EMA-VFI~\cite{emavfi2023}
 & 30.22 & 0.9031 & 0.0487 & 0.0533 & 0.5438
 & 24.49 & 0.7646 & 0.0922 & 0.1007 & 0.9959
 & 21.02 & 0.6627 & 0.1719 & 0.1938 & 1.5353 \\
\midrule
\grouprow{16}{Diffusion-based}
LDMVFI~\cite{ldmvfi2024}
 & 28.77 & 0.8787 & 0.0565 & 0.0640 & 0.6950
 & 23.70 & 0.7411 & 0.1072 & 0.1228 & 1.2991
 & 20.47 & 0.6348 & 0.1830 & 0.2060 & 2.0141 \\
CBBD~\cite{cbbd2024}
 & 29.65 & 0.8944 & 0.0390 & 0.0457 & 0.6275
 & 23.87 & 0.7499 & 0.0840 & 0.1037 & 1.3145
 & 20.37 & 0.6405 & 0.1616 & 0.1903 & 2.1002 \\
MoMo~\cite{momo2025}
 & 29.70 & 0.8924 & \textbf{0.0365} & 0.0392 & 0.5554
 & 24.10 & 0.7502 & 0.0743 & 0.0802 & 1.0030
 & 20.70 & 0.6451 & 0.1347 & 0.1481 & 1.5137 \\
EDEN~\cite{eden2025}
 & 29.44 & 0.8905 & 0.0453 & 0.0478 & 0.5713
 & 24.43 & 0.7628 & 0.0939 & 0.0994 & 1.0423
 & 21.24 & 0.6666 & 0.1699 & 0.1807 & 1.6710 \\
TLB-VFI~\cite{tlbvfi2025}
 & 29.58 & 0.8941 & 0.0373 & 0.0436 & 0.6310
 & 23.75 & 0.7467 & 0.0811 & 0.1004 & 1.3140
 & 20.58 & 0.6486 & 0.1471 & 0.1654 & 1.6808 \\
SPEED~\cite{speed2026}
 & 28.39 & 0.8764 & 0.0538 & 0.0552 & 0.5562
 & 24.06 & 0.7556 & 0.0941 & 0.0976 & 0.9861
 & 20.98 & 0.6605 & 0.1584 & 0.1670 & 1.4789 \\
LDF-VFI~\cite{ldfvfi2026}
 & 28.37 & 0.8775 & 0.0637 & 0.0660 & \underline{0.4886}
 & 24.26 & 0.7658 & 0.0920 & 0.0956 & \underline{0.8834}
 & 20.80 & 0.6520 & 0.1492 & 0.1572 & \underline{1.4186} \\
\oursrow
\textbf{\method}
 & \textbf{33.01} & \textbf{0.9478} & \underline{0.0367} & \textbf{0.0385} & \textbf{0.3280}
 & \textbf{25.84} & \textbf{0.8010} & \underline{0.0740} & \textbf{0.0766} & \textbf{0.7898}
 & \textbf{21.67} & \underline{0.6764} & \textbf{0.1304} & \textbf{0.1360} & \textbf{1.3559} \\
\bottomrule
\end{tabular}}
\end{table*}
\begin{table*}[!t]
\centering
\caption{\textbf{Quantitative comparison on Adobe240.}
Evaluation on all 17 test sequences at native $1280\times720$.
\textbf{Bold} and \underline{underline} indicate the best and second-best results.}
\label{tab:supp_adobe}
\setlength{\tabcolsep}{2.5pt}
\renewcommand{\arraystretch}{1.1}
\resizebox{\textwidth}{!}{%
\begin{tabular}{lccccccccccccccc}
\toprule
\multirow{2}{*}{\textbf{Method}}
 & \multicolumn{5}{c}{\textbf{Adobe240 $\times4$}}
 & \multicolumn{5}{c}{\textbf{Adobe240 $\times8$}}
 & \multicolumn{5}{c}{\textbf{Adobe240 $\times16$}} \\
\cmidrule(lr){2-6}\cmidrule(lr){7-11}\cmidrule(lr){12-16}
 & PSNR$\uparrow$ & SSIM$\uparrow$ & LPIPS$\downarrow$
 & \textls[-45]{FloLPIPS}$\downarrow$ & tOF$\downarrow$
 & PSNR$\uparrow$ & SSIM$\uparrow$ & LPIPS$\downarrow$
 & \textls[-45]{FloLPIPS}$\downarrow$ & tOF$\downarrow$
 & PSNR$\uparrow$ & SSIM$\uparrow$ & LPIPS$\downarrow$
 & \textls[-45]{FloLPIPS}$\downarrow$ & tOF$\downarrow$ \\
\midrule
\grouprow{16}{Non-diffusion}
FILM~\cite{film2022}
 & 34.89 & 0.9730 & \underline{0.0225} & 0.0248 & 1.0921
 & 31.57 & 0.9460 & 0.0326 & 0.0366 & 1.1693
 & 26.07 & 0.8592 & 0.0629 & 0.0734 & 1.3693 \\
RIFE~\cite{rife2022}
 & 34.00 & 0.9685 & 0.0256 & 0.0277 & 1.1484
 & 30.93 & 0.9421 & 0.0348 & 0.0384 & 1.2311
 & 25.74 & 0.8575 & 0.0624 & 0.0699 & 1.4789 \\
AMT~\cite{amt2023}
 & 34.99 & 0.9738 & 0.0308 & 0.0360 & 1.1750
 & 31.63 & 0.9473 & 0.0437 & 0.0523 & 1.2454
 & 26.16 & 0.8630 & 0.0838 & 0.1051 & 1.5145 \\
GIMM-VFI~\cite{gimmvfi2024}
 & \underline{35.10} & \underline{0.9746} & 0.0332 & 0.0377 & 1.0876
 & \underline{31.93} & \underline{0.9506} & 0.0475 & 0.0536 & \underline{1.1574}
 & \underline{26.72} & \underline{0.8740} & 0.0896 & 0.0986 & \underline{1.2916} \\
InterpAny~\cite{interpany2024}
 & 34.57 & 0.9717 & 0.0300 & 0.0367 & 1.1427
 & 31.26 & 0.9446 & 0.0438 & 0.0551 & 1.2211
 & 25.95 & 0.8616 & 0.0865 & 0.1056 & 1.4323 \\
BiM-VFI~\cite{bimvfi2025}
 & 34.88 & 0.9730 & 0.0253 & 0.0282 & 1.1210
 & 31.52 & 0.9454 & 0.0349 & 0.0392 & 1.2054
 & 26.13 & 0.8598 & 0.0632 & 0.0708 & 1.3493 \\
VTinker~\cite{vtinker2026}
 & 34.57 & 0.9719 & \underline{0.0225} & 0.0249 & 1.1082
 & 31.39 & 0.9451 & 0.0323 & \underline{0.0358} & 1.1775
 & 26.18 & 0.8607 & \underline{0.0600} & \underline{0.0675} & 1.3280 \\
EMA-VFI~\cite{emavfi2023}
 & 34.92 & 0.9735 & 0.0311 & 0.0364 & 1.1053
 & 31.76 & 0.9483 & 0.0421 & 0.0490 & 1.1714
 & 26.39 & 0.8664 & 0.0770 & 0.0909 & 1.3411 \\
\midrule
\grouprow{16}{Diffusion-based}
LDMVFI~\cite{ldmvfi2024}
 & 34.00 & 0.9671 & 0.0273 & 0.0311 & 1.1343
 & 30.42 & 0.9327 & 0.0445 & 0.0524 & 1.2819
 & 25.26 & 0.8420 & 0.0856 & 0.0995 & 1.5250 \\
CBBD~\cite{cbbd2024}
 & 34.54 & 0.9711 & 0.0235 & 0.0267 & 1.1121
 & 31.03 & 0.9408 & 0.0366 & 0.0456 & 1.2806
 & 25.46 & 0.8495 & 0.0746 & 0.0919 & 1.5791 \\
MoMo~\cite{momo2025}
 & 34.69 & 0.9718 & \textbf{0.0219} & \textbf{0.0242} & \underline{1.0868}
 & 31.41 & 0.9444 & \underline{0.0320} & 0.0359 & 1.1657
 & 26.08 & 0.8584 & 0.0606 & 0.0691 & 1.3151 \\
EDEN~\cite{eden2025}
 & 34.64 & 0.9713 & 0.0231 & 0.0251 & 1.1007
 & 31.40 & 0.9436 & 0.0359 & 0.0396 & 1.1899
 & 26.33 & 0.8625 & 0.0704 & 0.0775 & 1.3598 \\
TLB-VFI~\cite{tlbvfi2025}
 & 34.47 & 0.9712 & \underline{0.0225} & 0.0253 & 1.1234
 & 30.99 & 0.9411 & 0.0350 & 0.0436 & 1.2888
 & 25.69 & 0.8559 & 0.0688 & 0.0795 & 1.4191 \\
SPEED~\cite{speed2026}
 & 33.64 & 0.9664 & 0.0265 & 0.0281 & 1.0994
 & 30.82 & 0.9401 & 0.0383 & 0.0412 & 1.1731
 & 26.10 & 0.8606 & 0.0676 & 0.0733 & 1.3114 \\
LDF-VFI~\cite{ldfvfi2026}
 & 32.54 & 0.9606 & 0.0335 & 0.0369 & 1.1376
 & 30.76 & 0.9441 & 0.0406 & 0.0447 & 1.2109
 & 25.97 & 0.8660 & 0.0671 & 0.0726 & 1.3626 \\
\oursrow
\textbf{\method}
 & \textbf{35.40} & \textbf{0.9753} & 0.0228 & \underline{0.0247} & \textbf{1.0738}
 & \textbf{33.22} & \textbf{0.9610} & \textbf{0.0289} & \textbf{0.0318} & \textbf{1.1146}
 & \textbf{27.42} & \textbf{0.8840} & \textbf{0.0541} & \textbf{0.0577} & \textbf{1.2586} \\
\bottomrule
\end{tabular}}
\end{table*}

\FloatBarrier

\clearpage

\FloatBarrier

\clearpage

{
    \small
    \bibliographystyle{ieeenat_fullname}
    \bibliography{main}

@inproceedings{superslomo2018,
  author    = {Jiang, Huaizu and Sun, Deqing and Jampani, Varun and Yang, Ming-Hsuan and Learned-Miller, Erik and Kautz, Jan},
  title     = {{Super SloMo}: High Quality Estimation of Multiple Intermediate Frames for Video Interpolation},
  booktitle = {Proceedings of the IEEE/CVF Conference on Computer Vision and Pattern Recognition (CVPR)},
  pages     = {9000--9008},
  year      = {2018}
}

@inproceedings{vcii2018,
  author    = {Wu, Chao-Yuan and Singhal, Nayan and Krahenbuhl, Philipp},
  title     = {Video Compression through Image Interpolation},
  booktitle = {Proceedings of the European Conference on Computer Vision (ECCV)},
  pages     = {416--431},
  year      = {2018}
}

@inproceedings{softsplat2020,
  author    = {Niklaus, Simon and Liu, Feng},
  title     = {Softmax Splatting for Video Frame Interpolation},
  booktitle = {Proceedings of the IEEE/CVF Conference on Computer Vision and Pattern Recognition (CVPR)},
  pages     = {5437--5446},
  year      = {2020}
}

@inproceedings{qvi2019,
  author    = {Xu, Xiangyu and Siyao, Li and Sun, Wenxiu and Yin, Qian and Yang, Ming-Hsuan},
  title     = {Quadratic Video Interpolation},
  booktitle = {Advances in Neural Information Processing Systems (NeurIPS)},
  volume    = {32},
  pages     = {1647--1656},
  year      = {2019}
}

@inproceedings{cain2020,
  author    = {Choi, Myungsub and Kim, Heewon and Han, Bohyung and Xu, Ning and Lee, Kyoung Mu},
  title     = {Channel Attention Is All You Need for Video Frame Interpolation},
  booktitle = {Proceedings of the AAAI Conference on Artificial Intelligence (AAAI)},
  volume    = {34},
  number    = {07},
  pages     = {10663--10671},
  year      = {2020}
}

@inproceedings{raft2020,
  author    = {Teed, Zachary and Deng, Jia},
  title     = {{RAFT}: Recurrent All-Pairs Field Transforms for Optical Flow},
  booktitle = {European Conference on Computer Vision (ECCV)},
  pages     = {402--419},
  year      = {2020}
}

@inproceedings{rife2022,
  author    = {Huang, Zhewei and Zhang, Tianyuan and Heng, Wen and Shi, Boxin and Zhou, Shuchang},
  title     = {Real-Time Intermediate Flow Estimation for Video Frame Interpolation},
  booktitle = {European Conference on Computer Vision (ECCV)},
  pages     = {624--642},
  year      = {2022}
}

@inproceedings{film2022,
  author    = {Reda, Fitsum and Kontkanen, Janne and Tabellion, Eric and Sun, Deqing and Pantofaru, Caroline and Curless, Brian},
  title     = {{FILM}: Frame Interpolation for Large Motion},
  booktitle = {European Conference on Computer Vision (ECCV)},
  pages     = {250--266},
  year      = {2022}
}

@inproceedings{amt2023,
  author    = {Li, Zhen and Zhu, Zuo-Liang and Han, Ling-Hao and Hou, Qibin and Guo, Chun-Le and Cheng, Ming-Ming},
  title     = {{AMT}: All-Pairs Multi-Field Transforms for Efficient Frame Interpolation},
  booktitle = {Proceedings of the IEEE/CVF Conference on Computer Vision and Pattern Recognition (CVPR)},
  pages     = {9801--9810},
  year      = {2023}
}

@inproceedings{emavfi2023,
  author    = {Zhang, Guozhen and Zhu, Yuhan and Wang, Haonan and Chen, Youxin and Wu, Gangshan and Wang, Limin},
  title     = {Extracting Motion and Appearance via Inter-Frame Attention for Efficient Video Frame Interpolation},
  booktitle = {Proceedings of the IEEE/CVF Conference on Computer Vision and Pattern Recognition (CVPR)},
  pages     = {5682--5692},
  year      = {2023}
}

@inproceedings{interpany2024,
  author    = {Zhong, Zhihang and Krishnan, Gurunandan and Sun, Xiao and Qiao, Yu and Ma, Sizhuo and Wang, Jian},
  title     = {Clearer Frames, Anytime: Resolving Velocity Ambiguity in Video Frame Interpolation},
  booktitle = {European Conference on Computer Vision (ECCV)},
  pages     = {346--363},
  year      = {2024}
}

@inproceedings{gimmvfi2024,
  author    = {Guo, Zujin and Li, Wei and Loy, Chen Change},
  title     = {Generalizable Implicit Motion Modeling for Video Frame Interpolation},
  booktitle = {Advances in Neural Information Processing Systems (NeurIPS)},
  volume    = {37},
  pages     = {63747--63770},
  year      = {2024}
}

@inproceedings{bimvfi2025,
  author    = {Seo, Wonyong and Oh, Jihyong and Kim, Munchurl},
  title     = {{BiM-VFI}: Bidirectional Motion Field-Guided Frame Interpolation for Video with Non-uniform Motions},
  booktitle = {Proceedings of the IEEE/CVF Conference on Computer Vision and Pattern Recognition (CVPR)},
  pages     = {7244--7253},
  year      = {2025}
}

@inproceedings{vtinker2026,
  author    = {Wu, Chenyang and Fu, Jiayi and Guo, Chun-Le and Han, Shuhao and Li, Chongyi},
  title     = {{VTinker}: Guided Flow Upsampling and Texture Mapping for High-Resolution Video Frame Interpolation},
  booktitle = {Proceedings of the AAAI Conference on Artificial Intelligence (AAAI)},
  volume    = {40},
  number    = {13},
  pages     = {10638--10645},
  year      = {2026}
}

@inproceedings{davis2016,
  author    = {Perazzi, Federico and Pont-Tuset, Jordi and McWilliams, Brian and Van Gool, Luc and Gross, Markus and Sorkine-Hornung, Alexander},
  title     = {A Benchmark Dataset and Evaluation Methodology for Video Object Segmentation},
  booktitle = {Proceedings of the IEEE/CVF Conference on Computer Vision and Pattern Recognition (CVPR)},
  pages     = {724--732},
  year      = {2016}
}

@inproceedings{lavib2024,
  author    = {Stergiou, Alexandros},
  title     = {{LAVIB}: A Large-scale Video Interpolation Benchmark},
  booktitle = {Advances in Neural Information Processing Systems (NeurIPS)},
  volume    = {37},
  pages     = {29091--29105},
  year      = {2024}
}

@inproceedings{ldmvfi2024,
  author    = {Danier, Duolikun and Zhang, Fan and Bull, David},
  title     = {{LDMVFI}: Video Frame Interpolation with Latent Diffusion Models},
  booktitle = {Proceedings of the AAAI Conference on Artificial Intelligence (AAAI)},
  volume    = {38},
  number    = {2},
  pages     = {1472--1480},
  year      = {2024},
  doi       = {10.1609/aaai.v38i2.27912}
}

@inproceedings{madiff2024,
  author    = {Huang, Zhilin and Yu, Yijie and Yang, Ling and Qin, Chujun and Zheng, Bing and Zheng, Xiawu and Zhou, Zikun and Wang, Yaowei and Yang, Wenming},
  title     = {Motion-aware Latent Diffusion Models for Video Frame Interpolation},
  booktitle = {Proceedings of the ACM International Conference on Multimedia (ACM MM)},
  pages     = {1043--1052},
  year      = {2024},
  doi       = {10.1145/3664647.3680846}
}

@inproceedings{cbbd2024,
  author    = {Lyu, Zonglin and Li, Ming and Jiao, Jianbo and Chen, Chen},
  title     = {Frame Interpolation with Consecutive {B}rownian Bridge Diffusion},
  booktitle = {Proceedings of the ACM International Conference on Multimedia (ACM MM)},
  pages     = {3449--3458},
  year      = {2024},
  doi       = {10.1145/3664647.3680961}
}

@inproceedings{momo2025,
  author    = {Lew, Jaihyun and Choi, Jooyoung and Shin, Chaehun and Jung, Dahuin and Yoon, Sungroh},
  title     = {Disentangled Motion Modeling for Video Frame Interpolation},
  booktitle = {Proceedings of the AAAI Conference on Artificial Intelligence (AAAI)},
  volume    = {39},
  number    = {5},
  pages     = {4607--4615},
  year      = {2025},
  doi       = {10.1609/aaai.v39i5.32486}
}

@inproceedings{hfd2025,
  author    = {Hai, Yang and Wang, Guo and Su, Tan and Jiang, Wenjie and Hu, Yinlin},
  title     = {Hierarchical Flow Diffusion for Efficient Frame Interpolation},
  booktitle = {Proceedings of the IEEE/CVF Conference on Computer Vision and Pattern Recognition (CVPR)},
  pages     = {22943--22952},
  year      = {2025}
}

@inproceedings{eden2025,
  author    = {Zhang, Zihao and Chen, Haoran and Zhao, Haoyu and Lu, Guansong and Fu, Yanwei and Xu, Hang and Wu, Zuxuan},
  title     = {{EDEN}: Enhanced Diffusion for High-quality Large-motion Video Frame Interpolation},
  booktitle = {Proceedings of the IEEE/CVF Conference on Computer Vision and Pattern Recognition (CVPR)},
  pages     = {2105--2115},
  year      = {2025}
}

@inproceedings{tlbvfi2025,
  author    = {Lyu, Zonglin and Chen, Chen},
  title     = {{TLB-VFI}: Temporal-Aware Latent {B}rownian Bridge Diffusion for Video Frame Interpolation},
  booktitle = {Proceedings of the IEEE/CVF International Conference on Computer Vision (ICCV)},
  pages     = {16260--16269},
  year      = {2025}
}

@inproceedings{speed2026,
  author    = {Zhang, Zihao and Zhao, Haoyu and Yang, Siqian and Wu, Yidi and Jiang, Yudong and Wu, Zuxuan},
  title     = {{SPEED}: One-Step Pixel Diffusion for High-quality Video Frame Interpolation},
  booktitle = {Proceedings of the ACM International Conference on Multimedia (ACM MM)},
  year      = {2026}
}

@inproceedings{ldfvfi2026,
  author    = {Peng, Xinyu and Li, Han and Huang, Yuyang and Zheng, Ziyang and Wang, Yaoming and Chen, Xin and Dai, Wenrui and Li, Chenglin and Zou, Junni and Xiong, Hongkai},
  title     = {Towards Holistic Modeling for Video Frame Interpolation with Auto-regressive Diffusion Transformers},
  booktitle = {Proceedings of the IEEE/CVF Conference on Computer Vision and Pattern Recognition (CVPR)},
  year      = {2026}
}

@inproceedings{speedvfi2026,
  author    = {Ding, Ganggui and Xu, Xiaogang and Chen, Hao and Shen, Chunhua},
  title     = {{SpeedVFI}: One-step Diffusion for Efficient Video Frame Interpolation},
  booktitle = {Proceedings of the International Conference on Machine Learning (ICML)},
  series    = {Proceedings of Machine Learning Research},
  volume    = {306},
  pages     = {25263--25281},
  publisher = {PMLR},
  year      = {2026}
}

@inproceedings{rdvfi2026,
  author    = {Ma, Yongrui and Zhao, Shijie and Yao, Mingde and Li, Junlin and Zhang, Li and Liu, Xiaohong and Dou, Qi and Gu, Jinwei and Xue, Tianfan},
  title     = {Realtime Video Frame Interpolation using One-Step Diffusion Sampling},
  booktitle = {International Conference on Learning Representations (ICLR)},
  year      = {2026}
}

@inproceedings{snmvfi2026,
  author = {Jeong, Jisoo and Cai, Hong and Lin, Jamie Menjay
            and Ackermann, Hanno and Sim, Hyeonjun
            and Zhu, Yinhao and Shi, Yunxiao and Porikli, Fatih},
  title = {{SNM-VFI}: Symmetric Nonlinear Motion-Guided
           Generative Video Frame Interpolation},
  booktitle = {European Conference on Computer Vision Workshops (ECCVW)},
  year = {2026},
  url = {https://arxiv.org/abs/2608.13460}
}

@article{mog2025,
  author  = {Zhang, Guozhen and Zhu, Yuhan and Cui, Yutao and Zhao, Xiaotong and Ma, Kai and Wang, Limin},
  title   = {Motion-Aware Generative Frame Interpolation},
  journal = {arXiv preprint arXiv:2501.03699},
  year    = {2025}
}

@inproceedings{flowmatching2023,
  author    = {Lipman, Yaron and Chen, Ricky T. Q. and Ben-Hamu, Heli and Nickel, Maximilian and Le, Matt},
  title     = {Flow Matching for Generative Modeling},
  booktitle = {International Conference on Learning Representations (ICLR)},
  year      = {2023}
}

@inproceedings{rectifiedflow2023,
  author    = {Liu, Xingchao and Gong, Chengyue and Liu, Qiang},
  title     = {Flow Straight and Fast: Learning to Generate and Transfer Data with Rectified Flow},
  booktitle = {International Conference on Learning Representations (ICLR)},
  year      = {2023}
}

@inproceedings{vsd2023,
  author    = {Wang, Zhengyi and Lu, Cheng and Wang, Yikai and Bao, Fan and Li, Chongxuan and Su, Hang and Zhu, Jun},
  title     = {{ProlificDreamer}: High-Fidelity and Diverse Text-to-{3D} Generation with Variational Score Distillation},
  booktitle = {Advances in Neural Information Processing Systems (NeurIPS)},
  volume    = {36},
  pages     = {8406--8441},
  year      = {2023}
}

@inproceedings{dmd2024,
  author    = {Yin, Tianwei and Gharbi, Micha{\"e}l and Zhang, Richard and Shechtman, Eli and Durand, Fr{\'e}do and Freeman, William T. and Park, Taesung},
  title     = {One-step Diffusion with Distribution Matching Distillation},
  booktitle = {Proceedings of the IEEE/CVF Conference on Computer Vision and Pattern Recognition (CVPR)},
  pages     = {6613--6623},
  year      = {2024}
}

@inproceedings{osediff2024,
  author    = {Wu, Rongyuan and Sun, Lingchen and Ma, Zhiyuan and Zhang, Lei},
  title     = {One-Step Effective Diffusion Network for Real-World Image Super-Resolution},
  booktitle = {Advances in Neural Information Processing Systems (NeurIPS)},
  volume    = {37},
  pages     = {92529--92553},
  year      = {2024}
}

@inproceedings{seedvr22025,
  author    = {Wang, Jianyi and Lin, Shanchuan and Lin, Zhijie and Ren, Yuxi and Wei, Meng and Yue, Zongsheng and Zhou, Shangchen and Chen, Hao and Zhao, Yang and Yang, Ceyuan and Xiao, Xuefeng and Loy, Chen Change and Jiang, Lu},
  title     = {{SeedVR2}: One-Step Video Restoration via Diffusion Adversarial Post-Training},
  booktitle = {International Conference on Learning Representations (ICLR)},
  pages     = {40987--41009},
  year      = {2026}
}

@inproceedings{dove2025,
  author    = {Chen, Zheng and Zou, Zichen and Zhang, Kewei and Su, Xiongfei and Yuan, Xin and Guo, Yong and Zhang, Yulun},
  title     = {{DOVE}: Efficient One-Step Diffusion Model for Real-World Video Super-Resolution},
  booktitle = {Advances in Neural Information Processing Systems (NeurIPS)},
  volume    = {38},
  pages     = {85218--85237},
  year      = {2025}
}

@inproceedings{pssr2026,
  author    = {Wu, Aiqiu and Qiu, Zhaofan and Yao, Ting and Mei, Tao},
  title     = {{PS-SR}: Pseudo-Single-Step Video Super-Resolution via Speculative Diffusion},
  booktitle = {Proceedings of the IEEE/CVF Conference on Computer Vision and Pattern Recognition (CVPR)},
  pages     = {38218--38227},
  year      = {2026}
}

@article{wan2025,
      title={Wan: Open and Advanced Large-Scale Video Generative Models}, 
      author={{Team Wan} and Ang Wang and Baole Ai and Bin Wen and Chaojie Mao and Chen-Wei Xie and Di Chen and Feiwu Yu and Haiming Zhao and Jianxiao Yang and Jianyuan Zeng and Jiayu Wang and Jingfeng Zhang and Jingren Zhou and Jinkai Wang and Jixuan Chen and Kai Zhu and Kang Zhao and Keyu Yan and Lianghua Huang and Mengyang Feng and Ningyi Zhang and Pandeng Li and Pingyu Wu and Ruihang Chu and Ruili Feng and Shiwei Zhang and Siyang Sun and Tao Fang and Tianxing Wang and Tianyi Gui and Tingyu Weng and Tong Shen and Wei Lin and Wei Wang and Wei Wang and Wenmeng Zhou and Wente Wang and Wenting Shen and Wenyuan Yu and Xianzhong Shi and Xiaoming Huang and Xin Xu and Yan Kou and Yangyu Lv and Yifei Li and Yijing Liu and Yiming Wang and Yingya Zhang and Yitong Huang and Yong Li and You Wu and Yu Liu and Yulin Pan and Yun Zheng and Yuntao Hong and Yupeng Shi and Yutong Feng and Zeyinzi Jiang and Zhen Han and Zhi-Fan Wu and Ziyu Liu},
      journal = {arXiv preprint arXiv:2503.20314},
      year={2025}
}

@inproceedings{lora2022,
  author    = {Hu, Edward and Shen, Yelong and Wallis, Phillip and Allen-Zhu, Zeyuan and Li, Yuanzhi and Wang, Shean and Wang, Lu and Chen, Weizhu},
  title     = {{LoRA}: Low-Rank Adaptation of Large Language Models},
  booktitle = {International Conference on Learning Representations (ICLR)},
  year      = {2022}
}

@inproceedings{lpips2018,
  author    = {Zhang, Richard and Isola, Phillip and Efros, Alexei A. and Shechtman, Eli and Wang, Oliver},
  title     = {The Unreasonable Effectiveness of Deep Features as a Perceptual Metric},
  booktitle = {Proceedings of the IEEE/CVF Conference on Computer Vision and Pattern Recognition (CVPR)},
  pages     = {586--595},
  year      = {2018}
}

@inproceedings{unet2015,
  author    = {Ronneberger, Olaf and Fischer, Philipp and Brox, Thomas},
  title     = {{U-Net}: Convolutional Networks for Biomedical Image Segmentation},
  booktitle = {Medical Image Computing and Computer-Assisted Intervention (MICCAI)},
  pages     = {234--241},
  year      = {2015},
  doi       = {10.1007/978-3-319-24574-4_28}
}

@inproceedings{flolpips2022,
  author    = {Danier, Duolikun and Zhang, Fan and Bull, David},
  title     = {{FloLPIPS}: A Bespoke Video Quality Metric for Frame Interpolation},
  booktitle = {Picture Coding Symposium (PCS)},
  pages     = {283--287},
  year      = {2022},
  doi       = {10.1109/PCS56426.2022.10018062}
}

@article{fvd2018,
  author  = {Unterthiner, Thomas and van Steenkiste, Sjoerd and Kurach, Karol and Marinier, Raphael and Michalski, Marcin and Gelly, Sylvain},
  title   = {Towards Accurate Generative Models of Video: A New Metric \& Challenges},
  journal = {arXiv preprint arXiv:1812.01717},
  year    = {2018}
}

@article{ssim2004,
  author  = {Wang, Zhou and Bovik, Alan Conrad and Sheikh, Hamid Rahim and Simoncelli, Eero P.},
  title   = {Image Quality Assessment: From Error Visibility to Structural Similarity},
  journal = {IEEE Transactions on Image Processing},
  volume  = {13},
  number  = {4},
  pages   = {600--612},
  year    = {2004},
  doi     = {10.1109/TIP.2003.819861}
}

@article{tecogan2020,
  author  = {Chu, Mengyu and Xie, You and Mayer, Jonas and Leal-Taix\'{e}, Laura and Thuerey, Nils},
  title   = {Learning Temporal Coherence via Self-Supervision for {GAN}-based Video Generation},
  journal = {ACM Transactions on Graphics},
  volume  = {39},
  number  = {4},
  year    = {2020},
  doi     = {10.1145/3386569.3392457}
}

@inproceedings{abme2021,
        author    = {Park, Junheum and Lee, Chul and Kim, Chang-Su}, 
        title     = {Asymmetric Bilateral Motion Estimation for Video Frame Interpolation}, 
        booktitle = {Proceedings of the IEEE/CVF International Conference on Computer Vision (ICCV)},
        year      = {2021}
    }

@inproceedings{reds2019,
  title     = {{NTIRE} 2019 Challenge on Video Deblurring and Super-Resolution: Dataset and Study},
  author    = {Nah, Seungjun and Baik, Sungyong and Hong, Seokil and Moon, Gyeongsik and Son, Sanghyun and Timofte, Radu and Lee, Kyoung Mu},
  booktitle = {Proceedings of the IEEE/CVF Conference on Computer Vision and Pattern Recognition Workshops},
  year      = {2019}
}

@inproceedings{adobe2402017,
  title     = {Deep Video Deblurring for Hand-Held Cameras},
  author    = {Su, Shuochen and Delbracio, Mauricio and Wang, Jue and Sapiro, Guillermo and Heidrich, Wolfgang and Wang, Oliver},
  booktitle = {Proceedings of the IEEE Conference on Computer Vision and Pattern Recognition},
  pages     = {1279--1288},
  year      = {2017}
}

@inproceedings{videoinr2022,
  title     = {{VideoINR}: Learning Video Implicit Neural Representation for Continuous Space-Time Super-Resolution},
  author    = {Chen, Zeyuan and Chen, Yinbo and Liu, Jingwen and Xu, Xingqian and Goel, Vidit and Wang, Zhangyang and Shi, Humphrey and Wang, Xiaolong},
  booktitle = {Proceedings of the IEEE/CVF Conference on Computer Vision and Pattern Recognition},
  pages     = {2047--2057},
  year      = {2022}
}
}

\end{document}